\documentclass{article} 
\usepackage[preprint]{colm2026_conference}

\usepackage{microtype}
\usepackage{hyperref}
\usepackage{url}
\usepackage{xspace}
\usepackage{soul}
\usepackage{booktabs}
\usepackage{adjustbox}
\usepackage{pdflscape}  
\usepackage{amsmath}
\usepackage{tcolorbox}
\tcbuselibrary{skins, breakable}
\usepackage{enumitem}
\usepackage{fontawesome5}  
\usepackage{xcolor}
\usepackage{amsthm}
\usepackage{hhline}

\usepackage{acronym}
\usepackage{hyperref}
\usepackage{url}
\usepackage{algorithmicx}
\usepackage{algorithm}
\usepackage{algpseudocode}
\usepackage{tabularx}
\usepackage{booktabs}
\usepackage{colortbl}

\usepackage[utf8]{inputenc} 
\usepackage[T1]{fontenc}    
\usepackage{hyperref}       
\usepackage{url}            
\usepackage{booktabs}       
\usepackage{amsfonts}       
\usepackage{nicefrac}       
\usepackage{microtype}      
\usepackage{xcolor}         

\usepackage{booktabs}

\usepackage{booktabs}
\usepackage{multirow}
\usepackage{makecell}
\usepackage{tabularx}
\usepackage{array}
\usepackage{caption}

\usepackage[utf8]{inputenc}
\usepackage{fancybox}

\usepackage{lipsum}
\usepackage{float}         
\usepackage{caption}       

\usepackage[font=small, skip=0.5ex]{caption}
\usepackage[skip=0.3em, indent]{parskip}

\usepackage{booktabs}
\usepackage{array}
\newcolumntype{L}[1]{>{\raggedright\arraybackslash}p{#1}}
\newcolumntype{C}[1]{>{\centering\arraybackslash}p{#1}}

\newcommand{\sysname}{A\emph{PP}A\xspace}

\usepackage{lineno}

\definecolor{darkblue}{rgb}{0, 0, 0.5}
\hypersetup{colorlinks=true, citecolor=darkblue, linkcolor=darkblue, urlcolor=darkblue}

\title{\sysname: Adaptive Preference Pluralistic Alignment for Fair Federated RLHF of LLMs}

\author{Mahmoud Srewa, Tianyu Zhao, \& Salma Elmalaki \\
Department of Electrical Engineering and Computer Science\\
University of California, Irvine\\
Irvine, CA 92697, USA \\
\texttt{\{msrewa, tzhao15, salma.elmalaki\}@uci.edu} 
}

\acrodef{llms}[LLMs]{Large Language Models}
\acrodef{llm}[LLM]{Large Language Model}
\acrodef{fl}[FL]{Federated Learning}
\acrodef{gpo}[GPO]{Group Preference Optimization}
\acrodef{rlhf}[RLHF]{Reinforcement Learning from Human Feedback}
\acrodef{dpo}[DPO]{Direct Preference Optimization}
\acrodef{ppo}[PPO]{Proximal Policy Optimization}
\acrodef{fedrlhf}[FedRLHF]{Federated Reinforcement Learning From Human Feedback}
\acrodef{sft}[SFT]{Supervised Fine-Tuning}
\acrodef{qa}[Q/A]{Question Answering}
\acrodef{gpo}[GPO]{Group Preference Optimization}
\acrodef{grpo}[GRPO]{Group Robust Preference Optimization}
\acrodef{dpa}[DPA]{Distributional Preference Alignment}
\acrodef{opa}[OPA]{Ordinal Preference Alignment}
\begin{document}

\ifcolmsubmission
\linenumbers
\fi

\maketitle

\begin{abstract}

Aligning \ac{llms} with diverse human preferences requires pluralistic alignment, where a single model must respect the values of multiple distinct groups simultaneously. In \ac{fedrlhf}, these groups align a shared policy without centralizing preference data, which makes fair reward aggregation essential. Existing aggregation methods exhibit clear trade-offs: average-based aggregation systematically under-aligns worst-performing groups, while min aggregation prioritizes worst-group performance at the cost of overall alignment. We propose \textbf{\sysname}, an \textbf{A}daptive \textbf{P}reference \textbf{P}luralistic \textbf{A}lignment framework that dynamically reweights group-level rewards based on historical alignment rewards. Our approach prioritizes under-aligned groups without degrading well-aligned ones, while requiring no access to raw preference data. Integrated into a \ac{ppo}-based \ac{fedrlhf} pipeline and evaluated on GLOBALQA and OQA across three model families (Gemma-2-2B, Llama-3.2-3B, Qwen3-0.6B), \sysname achieves strong fairness–alignment trade-offs, improving worst-group alignment by up to $28\%$ over average aggregation while maintaining higher overall alignment than min aggregation across most configurations.

\end{abstract}

\section{Introduction}

\ac{llms} have shown strong capabilities across a wide range of tasks, from open-ended question answering to complex reasoning and creative generation~\cite{intro1,intro11}. Their real-world utility, however, depends critically on alignment with the diverse values, opinions, and social norms of their users~\cite{intro2}. This alignment challenge is fundamentally pluralistic: human societies consist of numerous distinct groups whose preferences can significantly diverge along demographic, cultural, and
geographic lines. A model that optimizes for a single aggregated preference distribution risks marginalizing minority groups and reinforcing majority bias~\cite{intro22}.

Prior work has explored prompt engineering and few-shot in-context learning as lightweight approaches to group preference alignment, but these methods do not reliably serve underrepresented groups at scale~\cite{intro3-prompt-no-2, intro3-prompt-no-3}. Gradient-based adaptation offers stronger alignment: \ac{sft} and per-group reward models can capture group-specific preferences more faithfully but require centralizing sensitive preference data and incur prohibitive computational costs when scaled to many groups~\cite{gpo, maxmin}. \ac{fl} offers a natural solution, enabling groups to contribute to a shared aligned policy without exposing their raw preference data \cite{fedsurvey2}.

\ac{fedrlhf} combines \ac{rlhf} with \ac{fl} so that heterogeneous groups can align on a shared policy without centralizing raw preference data, addressing both privacy and diversity limitations of centralized alignment \cite{fedsurvey,llmfed-2}. One instantiation of this paradigm forgoes local model training entirely: a server-hosted policy generates rollouts distributed to participating groups, each of which locally evaluates these rollouts and returns group-level rewards for these rollouts to the server for policy optimization, preserving data privacy while avoiding the communication overhead of exchanging model parameters \cite{srewa2025pluralllm}. However, the \ac{fl} setting introduces a critical and underexplored challenge: how to aggregate diverse and potentially conflicting group-level rewards in a way that is both effective and fair. Average aggregation treats all groups equally at each step, systematically under-aligning the worst-performing groups and embedding majority bias into the final policy~\cite{fairness-1, fairness-2, maxmin}. Min aggregation, while motivated by worst-case fairness, prioritizes the least-aligned group at the cost of overall alignment. Neither strategy jointly optimizes average and worst-group alignment across training.

We propose \textbf{\sysname}, an \textbf{A}daptive \textbf{P}reference \textbf{P}luralistic \textbf{A}lignment framework for fair \ac{fedrlhf}. \sysname dynamically reweights group-level rewards based on historical alignment rewards, assigning higher weights to under-aligned groups while maintaining non-zero weights for groups that are already better aligned. The method operates entirely on group-level rewards, requiring no access to raw preference data and preserving the privacy guarantees of the federated setting. By continuously tracking each group's alignment history and adjusting weights accordingly, \sysname mitigates both the majority bias of average aggregation and the tendency of min aggregation to overemphasize the worst-performing group, leading to consistent improvements in both worst-group and average alignment across most evaluated configurations. We summarize our main contributions as follows:

\begin{enumerate}

\item We study the problem of pluralistic alignment for \ac{llms}, focusing on how to align with diverse human preferences while promoting fairness across heterogeneous groups.

\item We design a novel adaptive reward aggregation algorithm that dynamically reweights group-level rewards based on historical alignment rewards, assigning higher weights to under-aligned groups while maintaining non-zero weights for better-aligned ones, without requiring access to raw preference data.

\item We evaluate \sysname on two \ac{qa} tasks using GLOBALQA and OQA across three model families (Gemma-2-2B, Llama-3.2-3B, Qwen3-0.6B), demonstrating consistent improvements over average and min aggregation baselines across most configurations.

\end{enumerate}

\section{Related Work}
\label{sec:related}

Prompt-based approaches~\cite{intro3-prompt-no-3, globalopnion} steer \ac{llms} toward
group preferences without weight updates but show only marginal gains on challenging
opinion surveys~\cite{gpo, globalopnion}. Gradient-based methods such as
\ac{gpo}~\cite{gpo} and PluralLLM~\cite{srewa2025pluralllm} learn to predict group
preference distributions directly, with PluralLLM doing so in a privacy-preserving
\ac{fl} setting, yet neither integrates these predictors into a full \ac{rlhf} loop.
On the \ac{rlhf} side, MaxMin-RLHF~\cite{maxmin} and GRPO~\cite{ramesh2024group} improve
worst-group alignment but concentrate policy updates on the single worst-performing
group at each step, discarding contributions from other groups; both also require centralized access to all group data. \ac{fedrlhf} methods such as FedBiOT~\cite{llmfed} and FedRLHF~\cite{llmfed-2} keep data local but do not address multi-group reward
aggregation. \sysname bridges these gaps: it uses PluralLLM group-level rewards within a \ac{ppo} loop and introduces adaptive reward aggregation that assigns adaptive weights to all groups at every iteration, improving both average and worst-group alignment simultaneously. See Appendix~\ref{app:related} for extended discussion.

\section{Proposed Method}
\label{sec:method}

\begin{figure*}[!t]
\centering
\includegraphics[width=1\textwidth]{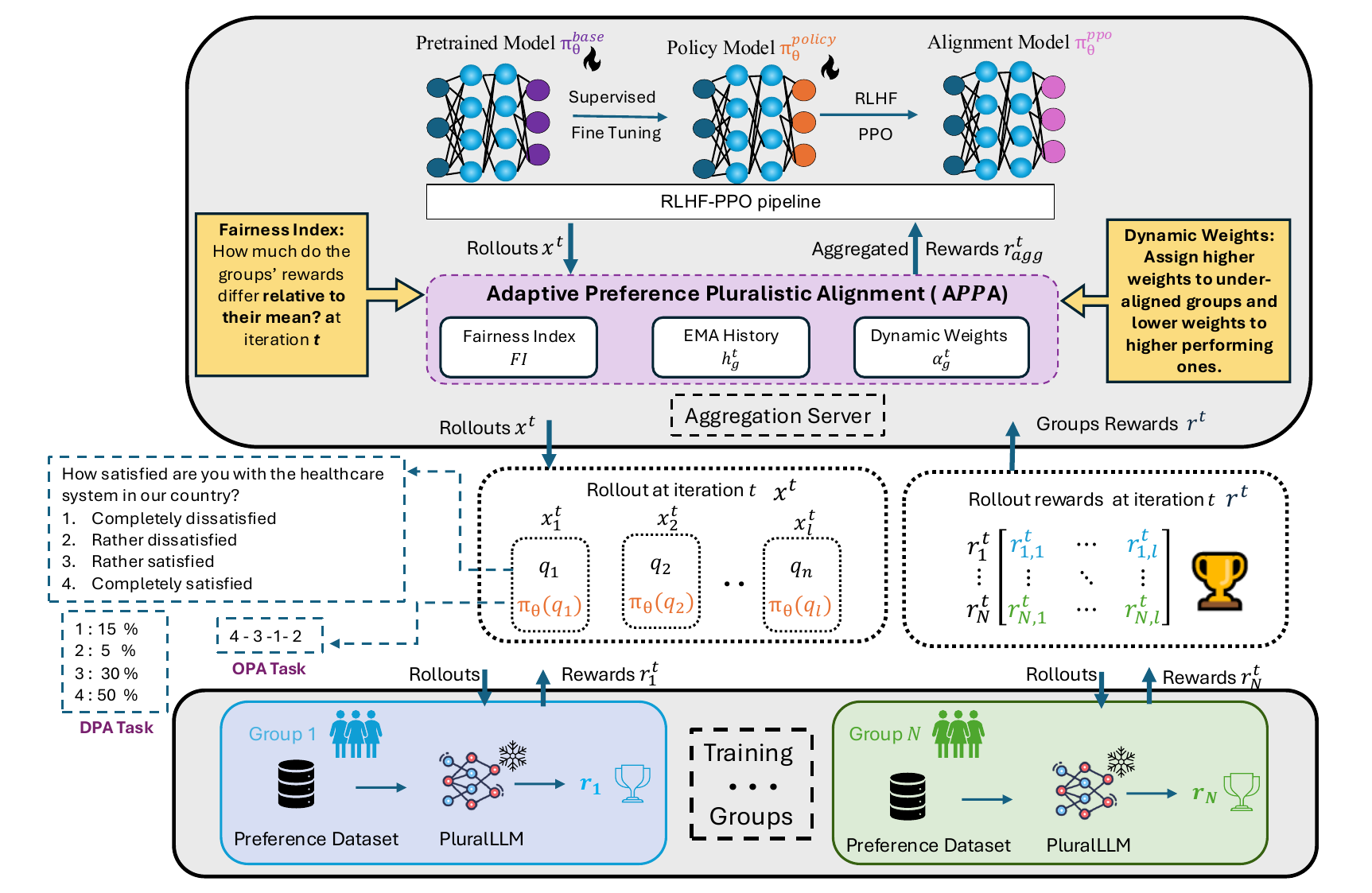}
\caption{
  Overview of \sysname: \ac{fedrlhf} for pluralistic alignment of group preferences.
  At each \ac{ppo} iteration, the server distributes rollouts to federated groups; each group scores responses using its local PluralLLM module and returns group-specific rewards. The server aggregates these rewards via the \sysname adaptive aggregation before updating the policy.
}
\label{fig:framework}
\end{figure*}

\subsection{System Setup and Training Groups}

We consider a \ac{fedrlhf} setting with $N$ training groups
$G_{\text{train}} = \{g_1, g_2, \ldots, g_N\}$,
where each group $g \in G_{\text{train}}$ corresponds to a distinct demographic or preference cluster
(e.g., a country in \textsc{GlobalQA} or a demographic group in OQA) and acts as a single federated client representing its members' aggregated preferences.  Each group maintains a private preference dataset  $D_g = \{(x_{g,j}, y_{g,j})\}_{j=1}^{m_g}$, 
where $m_g = |D_g|$ is the number of samples in group $g$, 
$x_{g,j}$ is the embedding of a query-response pair, and 
$y_{g,j}$ is the group's preference probability for that specific query-response pair. These datasets are never transmitted to the aggregation server. As illustrated in Figure~\ref{fig:framework}, an aggregation server initializes a base \ac{llm} $\pi^{\text{base}}_{\theta}$, performs \ac{sft} to obtain $\pi^{\text{policy}}_{\theta}$, and then iteratively updates the policy using group-level rewards from distributed evaluation.

\subsection{Distributed Reward Generation}

At iteration $t$, the server generates rollouts $X^t$ and broadcasts them to all groups. Each group $g$ uses PluralLLM~\cite{srewa2025pluralllm} (see Appendix~\ref{app:pluralllm}), a lightweight federated  few-shot preference predictor (used in a frozen manner), to output a probability distribution over the answer options for each question. For each rollout item $(q_j, y^{t,\text{policy}}_j) \in X^t$, where  $y^{t,\text{policy}}_j \sim \pi^{\text{policy}}_\theta(\cdot \mid q_j)$,  each group samples a small set of few-shot examples from its local dataset $D_g$ and conditions on these examples together with the rollout input to generate group-specific preference probabilities, which are then converted into per-item rewards. Let $r^t_{g,j}$ denote the reward assigned by group $g$ to item $(q_j, y^{t,\text{policy}}_j)$ at iteration $t$. For each item, the server receives the reward vector $\mathbf{r}^t_j = \{r^t_{g,j}\}_{g \in G_{\text{train}}}$.

\subsection{PPO Training Objective with Federated Reward}

Let $\mathcal{D}$ denote the global query distribution and $D_g$ the private preference
dataset of group $g$; these are related at the query level, though the server never
accesses any $D_g$ directly.

Standard PPO-RLHF optimizes a scalar reward subject to KL regularization. In
our pluralistic setting, we replace the centralized reward with an adaptive
aggregated reward $\mathrm{Agg}_{\boldsymbol{\alpha}^t}(\mathbf{r}^t_j)$(defined in Sec.~\ref{sec:adaptive-alpha}). Concretely, for each rollout item $q_j \in X^t$ we compute $\mathrm{Agg}_{\boldsymbol{\alpha}^t}(\mathbf{r}^t_j)$ from the received per-group rewards and use this scalar as the per-item PPO reward:
\begin{equation}
\label{eq:ppo-federated}
\max_{\pi_\theta}\;
\mathbb{E}_{x \sim \mathcal{D},\, y \sim \pi_\theta(\cdot|x)}
\!\left[\mathrm{Agg}_{\boldsymbol{\alpha}^t}\!\left(\mathbf{r}^t_j\right)-\beta\,
  \mathrm{KL}\!\left[\pi_\theta(\cdot|x)\,\|\,\pi_{\text{ref}}(\cdot|x)\right]
\right].
\end{equation}
The key design question is how to compute $\mathrm{Agg}_{\boldsymbol{\alpha}^t}(\cdot)$ so as
to improve the fairness-alignment trade-off between average and worst-group
performance. The full PPO objective is given in Appendix~\ref{app:ppo};  Our key modification is to replace the centralized reward $r(x,y)$
with the aggregated per-item reward $\mathrm{Agg}_{\boldsymbol{\alpha}^t}(\mathbf{r}^t_j)$.
\subsection{Adaptive Alpha Aggregation}
\label{sec:adaptive-alpha}

\paragraph{Background: Alpha Aggregation.}
~\cite{park2024rlhf} introduce reward aggregation rules grounded in social
choice theory, parameterized by a scalar $\alpha \in \mathbb{R}$:
\begin{equation}
\label{eq:consensus-aggregation}
\mathrm{Agg}_\alpha(\mathbf{r})=
\begin{cases}
\dfrac{1}{\alpha}\log\!\left(\dfrac{1}{N}\sum_{i=1}^{N}\exp(\alpha r_i)\right) & \alpha \neq 0,\\[6pt]
\dfrac{1}{N}\sum_{i=1}^{N} r_i & \alpha = 0,
\end{cases}
\end{equation}
interpolating from $\min_i r_i$ ($\alpha\!\to\!-\infty$) to $\max_i r_i$
($\alpha\!\to\!+\infty$), with fairness-theoretic guarantees (monotonicity, symmetry,
translation independence, and Pigou-Dalton transfer). \cite{park2024rlhf} treats $\alpha$ as a fixed scalar chosen before training
and applied identically to all individuals throughout optimization.
In our \ac{fedrlhf} setting we instantiate $N = |G_{\text{train}}|$ groups,
each acting as a single federated client.
\vspace{-3pt}
\paragraph{Our Contribution: Group-Specific Adaptive Weights.}
We replace the single global $\alpha$ with \emph{group-specific, dynamically updated}
weights $\alpha_g^t$ that continuously upweight groups with lower accumulated alignment. We additionally compute a Fairness Index ($FI$, defined below) and apply a threshold to determine when adaptive weighting is used:
\vspace{-3pt}
\begin{equation}
\label{eq:adaptive-aggregation}
\mathrm{Agg}_{\boldsymbol{\alpha}^t}(\mathbf{r}^t_j) =
\begin{cases}
\displaystyle\frac{1}{|G_{\text{train}}|}\sum_{g \in G_{\text{train}}} r^t_{g,j}
  & \text{if } FI \ge \tau, \\[8pt]
\log\!\left(\dfrac{1}{|G_{\text{train}}|}
  \sum_{g \in G_{\text{train}}} \exp\!\left(\alpha_g^t \cdot r^t_{g,j}\right)\right)
  & \text{otherwise,}
\end{cases}
\end{equation}
where $\tau = 0.99$ is a fairness threshold set such that the adaptive
aggregation remains active for the large majority of training (see Appendix \ref{appendix:setup} for justification).
When cross-group rewards are already highly uniform ($FI \ge \tau$), a plain average prevents unnecessary adjustments that could destabilize training; when $FI < \tau$, the log-sum-exp with group-specific $\alpha_g^t$ steers the policy toward under performing groups.
Unlike \cite{park2024rlhf}, which uses a single scalar $\alpha$ and therefore
includes the scalar normalization factor $\tfrac{1}{\alpha}$, our formulation
uses group-specific adaptive weights $\alpha_g^t$. We therefore use a modified
log-sum-exp aggregator rather than a direct instantiation of the fixed-$\alpha$
formulation. Since $r^t_{g,j} \in [0,1]$ and $\alpha_g^t \in (0,1)$ with
$\sum_g \alpha_g^t = 1$, the resulting aggregated reward remains bounded.

\vspace{-2pt}

\paragraph{Fairness Index.}
To quantify within-iteration reward disparity, we define a Fairness Index
($FI$) based on the coefficient of variation ($CoV$) of per-group rewards for
each question $q_j$. The $CoV$ measures relative spread: a high $CoV$ indicates
that group rewards diverge substantially around their mean, while a $CoV$ near
zero indicates near-uniform reward across groups.
\vspace{-4pt}
\begin{equation}
\label{eq:fi}
FI \;=\; \frac{1}{|X^t|}\sum_{q_j \in X^t}
\frac{1}{1+\mathrm{CoV}^2(q_j)},
\qquad
\mathrm{CoV}(q_j) =
\frac{\sigma\!\left(\{r^t_{g,j}\}_{g \in G_{\text{train}}}\right)}
  {\mu\!\left(\{r^t_{g,j}\}_{g \in G_{\text{train}}}\right)}.
\end{equation}
$FI \in [0,1]$, where $FI=1$ indicates identical rewards across all groups and
$FI \to 0$ indicates increasing inter-group disparity. Edge cases such as
near-zero mean rewards and extreme $CoV$ values are handled via numerical
safeguards described in Appendix~\ref{appendix:fi-safeguards}.
\paragraph{Adaptive Weight Computation via Reverse Softmax.}
We maintain a \emph{historical alignment score} $h_g^t$ updated via exponential moving
average:
\vspace{-4pt}
\begin{equation}
\label{eq:ema}
h_g^t \;=\; \lambda \cdot h_g^{t-1} \;+\; (1-\lambda) \cdot \bar{r}^t_g,
\end{equation}
where $\bar{r}^t_g = \frac{1}{|X^t|}\sum_{q_j \in X^t} r^t_{g,j}$ and $\lambda = 0.8$.
The EMA smoothing prevents transient reward fluctuations from dominating the weight
update, providing a stable long-run picture of each group's historical alignment rewards.
Group weights are then computed via a \emph{reversed softmax}; at iteration $t$,
$\alpha_g^t$ is computed from the \emph{previous} EMA history $h_g^{t-1}$:
\vspace{-4pt}
\begin{equation}
\label{eq:reversed-softmax}
\alpha_g^t \;=\; \frac{\exp\!\left(\dfrac{1 - h_g^{t-1}}{T}\right)}
{\displaystyle\sum_{g' \in G_{\text{train}}}
  \exp\!\left(\dfrac{1 - h_{g'}^{t-1}}{T}\right)},
\qquad T = 0.1.
\end{equation}
Inverting the history ($1 - h_g^{t-1}$) ensures lower-performing groups receive higher
weights; the temperature $T=0.1$ sharpens the distribution to concentrate signal on
lagging groups while keeping all $\alpha_g^t > 0$.
\vspace{-2pt}

\paragraph{Comparison with~\cite{park2024rlhf}.}
Table~\ref{tab:contribution-diff} summarizes the key differences.
Our contribution is orthogonal: we embed alpha aggregation in a federated RLHF loop
with local PluralLLM reward predictors, replace the static scalar with adaptive
per-group weights, and introduce the Fairness Index with a threshold-based aggregation rule together yielding stronger fairness-alignment trade-offs than fixed-$\alpha$ baselines and hard minimax strategies across our evaluated settings.

\begin{table}[t]
\centering
\caption{Key differences between~\cite{park2024rlhf} and \sysname (ours).}
\label{tab:contribution-diff}
\small
\renewcommand{\arraystretch}{1.3}
\arrayrulecolor{gray!40}
\begin{tabularx}{\textwidth}{>{\bfseries}m{2.6cm} X X}
\arrayrulecolor{black}\hline\arrayrulecolor{gray!40}
\textbf{Property} & \textbf{~\cite{park2024rlhf}} & \textbf{\sysname (Ours)} \\
\arrayrulecolor{black}\hline\arrayrulecolor{gray!40}
Aggregation weight $\alpha$ &
Single fixed scalar (manual, same for all, never updated) &
Per-group $\alpha_g^t$: automatic, data-driven, updated each PPO step \\\hline
Reward source &
Centralized reward model &
Federated PluralLLM per group \\\hline
Data sharing &
Required &
Not required (privacy-preserving) \\\hline
Fairness-aware threshold &
\texttimes &
\checkmark\ ($FI$ threshold) \\\hline
History tracking &
\texttimes &
EMA over group alignment history \\\hline
Weight computation &
Manually swept ($\alpha \in \{-\infty,-1,0,1,+\infty\}$) &
Reversed softmax over EMA history (Eq.~\ref{eq:reversed-softmax}) \\\hline
Training paradigm &
Iterative PPO, fixed $\alpha$ &
Iterative PPO, $\alpha_g^t$ recomputed each iteration \\\hline
Empirical validation &
Text summarisation (1 task) &
GLOBALQA \& OQA (2 tasks, 3 models) \\
\arrayrulecolor{black}\hline
\end{tabularx}
\arrayrulecolor{black}
\end{table}


\section{Algorithm Details}
\label{app:algorithm}

Algorithm~\ref{alg:appa} summarizes the full procedure; formal PPO compatibility
and gradient bias analysis are in Appendix~\ref{app:theory}.

\begin{algorithm}[t]
\caption{\sysname: Adaptive Pluralistic Preference Alignment via Federated RLHF}
\label{alg:appa}
\begin{algorithmic}[1]
\Require
  Base LLM $\pi^{\text{base}}_\theta$;
  groups $G_{\text{train}} = \{g_1,\ldots,g_N\}$, each group $g$ with dataset $D_g$
  and a frozen PluralLLM;
  PPO hyperparameters $(\beta, \eta)$;
  $\lambda{=}0.8$, $T{=}0.1$, $\tau{=}0.99$.
\Ensure Aligned policy $\pi^{\text{policy}}_{\theta}$.
\State \textbf{Initialize:} SFT on $\pi^{\text{base}}_\theta \to \pi^{\text{policy}}_\theta$;
       set $h_g^0 \leftarrow 0\;\forall g$.
\For{iteration $t = 1, 2, \ldots, T_{\max}$}
  \State \textbf{Rollout:} Sample $X^t \sim \pi^{\text{policy}}_\theta$; broadcast to all groups.
  \For{each $g \in G_{\text{train}}$ \textbf{(in parallel)}}
    \State Compute $r^t_{g,j} \leftarrow \mathcal{R}\!\left(\pi_\theta(q_j),\;
       \hat{p}^{\text{PluralLLM}}_{g,j}\right)$; return $\{r^t_{g,j}\}$ to the server.
  \EndFor
    \State Compute $FI$ (Eq.~\eqref{eq:fi}), $\alpha_g^t$ (Eq.~\eqref{eq:reversed-softmax}),
      and $\{\mathrm{Agg}_{\boldsymbol{\alpha}^t}(\mathbf{r}^t_j)\}_{q_j \in X^t}$ (Eq.~\eqref{eq:adaptive-aggregation}).
  \State \textbf{PPO update:} Update $\pi_\theta$ via Eq.~\eqref{eq:ppo-federated}.
    \State \textbf{Update history:} $h_g^t \leftarrow \lambda\,h_g^{t-1} + (1{-}\lambda)\,\bar{r}^t_g\;\forall g$.
\EndFor
\State \Return $\pi^{\text{policy}}_\theta$.
\end{algorithmic}
\end{algorithm}

\section{Evaluation}
\label{sec:eval}

\subsection{Tasks and Dataset Construction}
\label{subsec:tasks}

\paragraph{Models and Datasets.}
We evaluate pluralistic alignment across two complementary tasks that probe different facets of group preference modeling, using two opinion datasets that differ in cultural scope and answer structure and three instruction-tuned model families ranging from 0.6B to 3B parameters. The datasets are \textbf{GLOBALQA} (Pew Research Global Attitudes~\cite{globalopnion}), which tests \emph{cross-national} pluralistic alignment with nominal answer spaces, and \textbf{OQA} (OpinionQA~\cite{intro3-prompt-no-3}), which tests \emph{intra-national} demographic alignment with ordinal preference structure. The evaluated models are Gemma-2-2B, Llama-3.2-3B, and Qwen3-0.6B (see Appendix~\ref{appendix:models}). Full dataset details are in Appendix~\ref{appendix:datasets}. For both datasets, we use an $80/20$ train--test split, and all reported results are evaluated on the held-out test set.

\paragraph{\ac{dpa}.}
For each survey question $q_j$ and group $g$, we obtain a
\emph{target preference distribution} 
$\hat{p}^{\text{PluralLLM}}_{g,j} \in \Delta^{K-1}$ over the $K$ answer options
from a frozen PluralLLM model~\cite{srewa2025pluralllm} (see Appendix~\ref{app:pluralllm}), which predicts the probability of each answer option being preferred by group $g$ for question $q_j$.
PluralLLM is trained to predict group-level response distributions from
survey data; we therefore treat $\hat{p}^{\text{PluralLLM}}_{g,j}$ as our proxy “ground truth” for group $g$ on question $q_j$.
The policy model $\pi_\theta$ is prompted to output a probability distribution over the same set of options (see Figure~\ref{fig:prob-prompt}).
Reward functions then measure how close the model’s predicted distribution
$y^{t,\text{policy}}_j$ is to $\hat{p}^{\text{PluralLLM}}_{g,j}$, using
Jensen–Shannon divergence (JS), Wasserstein distance (Was.), and cosine
similarity (Cos.). This task evaluates whether the model captures the full structure of group
preferences, not only which option is preferred but also how strongly each
option is supported.

\paragraph{\ac{opa}.}
The OPA task is derived from the same distributions used in \ac{dpa}.
For each group $g$, we obtain a ranking $\sigma_g(q_j)$ by sorting
$P_g(q_j)$ in descending order, requiring no additional annotation. The policy is prompted to produce a ranked list over the answer options
(see Figure~\ref{fig:rank-prompt}), and reward functions evaluate the
agreement between rankings using the Borda score (Bor.).

\paragraph{SFT Baseline and Majority Aggregation.}
The SFT model is fine-tuned on a single majority-aggregated label per
question (the most frequent response across all groups).
SFT therefore learns the modal preference rather than the full group
distribution, systematically under-serving minority groups, a degradation
that PPO with federated reward aggregation is designed to reverse.

\subsection{Evaluation Framework}
\label{subsec:framework}

We parse each response to extract prediction $y^{t,\text{policy}}_j$ and, for
each group $g \in G_{\text{train}}$, compute group-level reward $r^{t}_{g,j}$
by comparing against the group target $\hat{p}^{\text{PluralLLM}}_{g,j}$~\cite{srewa2025pluralllm}. We treat PluralLLM predictions $\hat{p}^{\text{PluralLLM}}_{g,j}$ as estimates of the empirical survey distributions $P_g(q_j)$, following~\citet{srewa2025pluralllm}. This is combined with a format score to form the final PPO reward signal (see Appendix~\ref{appendix:format}).

\subsubsection{Reward Metrics}
\label{subsubsec:metrics}

For DPA we compute Jensen–Shannon divergence (JS), Wasserstein distance (Was.), and cosine similarity (Cos.) rewards; for OPA we use the Borda score (Bor.).
All metrics are normalized to $[0,1]$; higher values indicate better
alignment. JS is the primary DPA metric for GLOBALQA and Wasserstein distance
for OQA, reflecting their nominal and ordinal answer structures respectively.
Full definitions and rationale are in Appendix~\ref{appendix:metrics}. To summarize performance across groups, we compute per-group alignment rewards by averaging
per-question rewards over the evaluation set, then report \emph{Avg AS} and \emph{Min AS} by
taking the mean and minimum across groups; formal definitions are in Appendix~\ref{appendix:as-def}.

\subsubsection{Fairness Index}

$FI$ quantifies reward variation across groups over the rollout, based on per-question rewards as defined in Equation~\eqref{eq:fi}, where fairness is first computed at the question level and then averaged across all questions in the rollout.
$FI \in [0,1]$; $FI = 1$ indicates identical rewards across all groups
(maximum fairness), while $FI \to 0$ indicates systematic between-group
disparity. All aggregate results are reported in
Tables~\ref{tab:combined-value-ppo-base-sft} and~\ref{tab:combined-order-ppo-base-sft}
and reflect model performance at the \emph{final PPO iteration}.
The high terminal $FI$ values for \textsc{\sysname} ($\geq 0.999$ on GLOBALQA
DPA) are a \emph{consequence} of successful alignment: early in training,
when inter-group disparity is high ($FI < \tau$), adaptive log-sum-exp
aggregation steers the policy toward under-aligned groups; as alignment
converges, $FI$ rises toward $\tau$ and aggregation falls back to a plain
average.

\subsubsection{Aggregation Strategies}

We compare three server-side reward aggregation schemes.
\textbf{Average} computes the mean reward across all groups, giving equal
weight to every group regardless of alignment level, which can mask
systematic under-service of under-aligned groups.
\textbf{Min} is a minimax strategy that optimizes exclusively through the
worst-performing group at each step, discarding rewards from all
others and risking stalled learning once the worst group improves.
\textbf{\sysname (Ours)} assigns a \emph{continuous adaptive weight}
$\alpha_g^t$ to every group based on historical alignment rewards
(Section~\ref{sec:adaptive-alpha}), prioritizing under-served groups while
ensuring all groups contribute to the update. When cross-group rewards are
already uniform ($FI \ge \tau = 0.99$), aggregation falls back to a plain
average to prevent over-correction.

\subsection{Quantitative Results: GLOBALQA Dataset}
\label{subsec:results-globalqa}

\subsubsection{Distributional Preference Alignment (DPA) on GLOBALQA}
\label{subsubsec:dpa-globalqa}

Table~\ref{tab:combined-value-ppo-base-sft} reports DPA results on GLOBALQA,
using JS as the main aggregation reward given GLOBALQA's nominal answer
structure; Was.\ and Cos.\ are logged and reported as secondary measures.

SFT moves the fairness needle very little: base models achieve JS FI of
$0.938$--$0.939$, and SFT leaves these essentially unchanged despite lifting
average alignment scores substantially for Gemma-2-2B and Llama-3.2-3B (about
$+0.10$--$+0.12$ in JS Avg AS), but only marginally for Qwen3-0.6B. Distributional reward feedback
is what drives fairness: PPO with any aggregation pushes FI above $0.969$,
and with \textsc{\sysname} or \textsc{Min} reaches $\geq 0.999$.

\textsc{\sysname} achieves the highest Avg AS \emph{and} Min AS simultaneously
across all three model families
(Gemma-2-2B: JS $0.861$/$0.843$, FI $0.9994$;
 Llama-3.2-3B: $0.848$/$0.834$, FI $0.9995$;
 Qwen3-0.6B: $0.824$/$0.809$, FI $0.9996$).
\textsc{Min} reaches comparable FI but lower Avg AS and Min AS; \textsc{Average}
is competitive on Gemma-2-2B and Llama-3.2-3B but performs markedly worse on
Qwen3-0.6B, with both lower Avg AS and Min AS.
\textsc{\sysname} is the only strategy that avoids trading one against the other.

\subsubsection{Ordinal Preference Alignment (OPA) on GLOBALQA}
\label{subsubsec:opc-globalqa}

Table~\ref{tab:combined-order-ppo-base-sft} reports OPA results on GLOBALQA,
using Borda as the main aggregation reward.

Ranking is a harder objective than distributional prediction: SFT Borda
scores of $0.339$--$0.461$ are well below the corresponding DPA Avg AS
($\approx 0.73$--$0.79$), and for Gemma-2-2B SFT \emph{worsens} ranking relative to base
(Borda Avg AS $0.461 \to 0.339$; Min AS $0.434 \to 0.329$), a known
consequence of majority-label fine-tuning misaligning minority group
rankings~\cite{intro3-prompt-no-2,intro3-prompt-no-3}. PPO reverses this
across all models.

\textsc{\sysname} leads on Avg AS and Min AS across all model families, and on FI for Gemma-2-2B and Llama-3.2-3B; for Qwen3-0.6B, \textsc{\sysname} matches \textsc{Min} on FI ($0.821$) while still outperforming it on both Avg AS and Min AS.
(Gemma-2-2B: FI $0.891$, Avg $0.511$, Min $0.461$;
 Llama-3.2-3B: FI $0.882$, Avg $0.536$, Min $0.487$;
 Qwen3-0.6B: Avg $0.442$, Min $0.429$).
\textsc{Min} recovers some worst-group performance but at a consistent cost
to Avg AS; \textsc{Average} sustains Avg AS but leaves FI and Min AS behind.

\subsection{Quantitative Results: OQA Dataset}
\label{subsec:results-oqa}

\subsubsection{Distributional Preference Alignment (DPA) on OQA}
\label{subsubsec:dpa-oqa}
Table~\ref{tab:combined-value-ppo-base-sft} reports DPA results on OQA. OQA groups are US demographic groups rather than countries, and answers are predominantly ordinal; accordingly, Wasserstein is used as the main PPO aggregation reward. JS and Cos.\ are logged as secondary measures.
Base model Avg AS (Was.) sits at $0.681$--$0.763$ across models, with Min AS
substantially lower ($0.561$--$0.658$), indicating high dispersion across
demographic groups. \textsc{\sysname} achieves the highest Avg AS and Min AS simultaneously for Gemma-2-2B and Llama-3.2-3B
(Gemma-2-2B: Was.\ $0.872$/$0.842$, FI $0.9942$; Llama-3.2-3B: Was.\ $0.872$/$0.841$, FI $0.9940$),
confirming that adaptive weighting generalizes robustly across model
capacities. \textsc{Average} and \textsc{Min} both fall short on at least one
key metric in the higher-capacity settings. Format scores remain high under
PPO and are often near-perfect; in particular, \textsc{\sysname} reaches a
format score of $1.000$ for Llama-3.2-3B and Qwen3-0.6B.
The one notable exception is Qwen3-0.6B on OQA DPA, where \textsc{Min}
achieves higher Avg AS ($0.823$) than \textsc{\sysname} ($0.780$);
we attribute this to the smaller model capacity limiting the policy's
responsiveness to the Wasserstein gradient signal, leaving insufficient
contrast for the adaptive weights to steer optimization toward under-aligned
demographic groups.

\subsubsection{Ordinal Preference Alignment (OPA) on OQA}
\label{subsubsec:opc-oqa}

Table~\ref{tab:combined-order-ppo-base-sft} reports OPA results on OQA, using Borda as the main aggregation reward.

PPO raises Borda FI by $+0.08$--$+0.11$ and Avg AS by $+0.03$--$+0.10$
over SFT across all models. \textsc{\sysname} leads on Borda FI across all three
models ($0.714$ / $0.721$ / $0.704$ for Gemma-2-2B / Llama-3.2-3B /
Qwen3-0.6B). For Llama-3.2-3B, \textsc{\sysname} leads on Avg AS ($0.499$) and achieves higher Fairness Index ($0.721$) than both baselines, while \textsc{Min} yields
lower Avg AS ($0.434$) despite a marginally higher Min AS ($0.294$ vs.\ $0.285$),
illustrating that adaptive weighting avoids the fairness-efficiency tension
at the mean level without fully sacrificing worst-group floor.

\subsection{Per-Group Alignment: Spider Plot Analysis}
\label{subsec:spider}

\begin{figure*}[t]
  \centering
  \includegraphics[width=\textwidth]{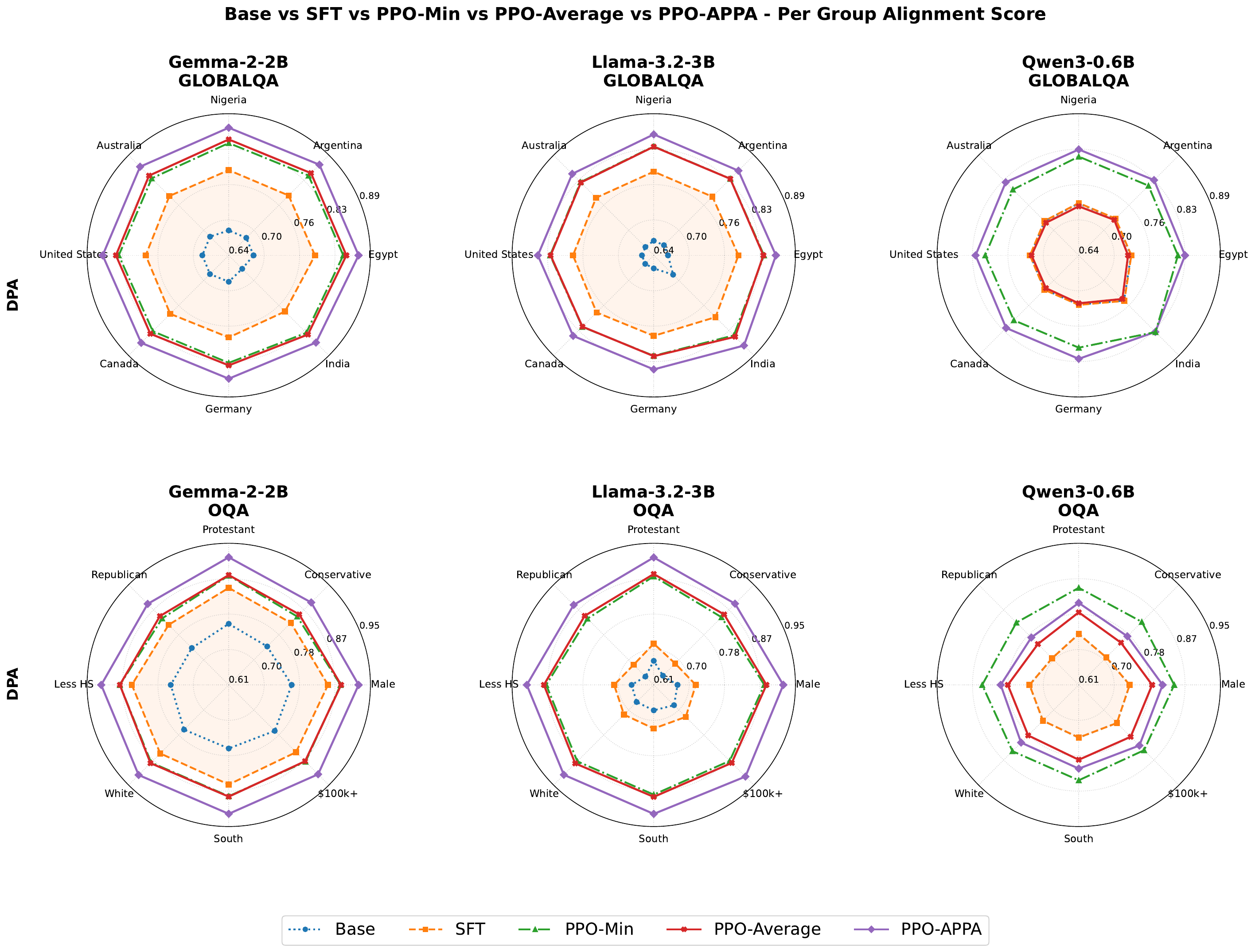}
\caption{%
    \textbf{Per-group alignment score comparison across selected
    demographic groups (Base vs.\ SFT vs.\ PPO-Min vs.\ PPO-Average vs.\
    PPO-\sysname).}
    Top row: GLOBALQA, eight countries with diverse opinions (JS metric).
    Bottom row: OQA, eight US demographic groups with diverse opinions
    (Wasserstein metric).
    A wider, more uniformly filled polygon indicates higher and more equitable
    alignment across groups.
}
  \label{fig:spider-plots}
\end{figure*}

Figure~\ref{fig:spider-plots} breaks aggregate numbers down to the per-group level. On GLOBALQA (Nigeria, Argentina, Egypt, Australia, United States,
Canada, Germany, India), PPO-\sysname scores land in $0.88$--$0.89$ across all
eight countries for Gemma-2-2B, with gains of $+0.19$--$+0.21$ over base.
The polygon is both larger and more circular than those of competing
strategies, indicating gains are spread rather than concentrated. Llama-3.2-3B
and Qwen3-0.6B show somewhat smaller absolute gains ($+0.09$--$+0.18$) with the same uniformity. No country regresses under PPO-\sysname relative to base.

On OQA (Protestant, Conservative, Male, \$100k+, South, White, Less HS,
Republican), absolute gains are larger ($+0.14$--$+0.16$ for Gemma-2-2B),
consistent with the ordinal reward signal being more discriminative. The
lowest-scoring group under PPO-\sysname (Male, $0.88$) still sits well above
its base score ($\approx 0.74$). For Gemma-2-2B and Llama-3.2-3B, PPO-Average produces a visibly less uniform polygon: some groups score higher than under PPO-\sysname, but others score lower, reflecting the absence of any mechanism to prevent dominant groups from capturing the gradient signal.

\subsection{Fairness--Alignment Trade-off Analysis}
\label{subsec:tradeoff}

\begin{figure*}[t]
  \centering
  \includegraphics[width=\textwidth]{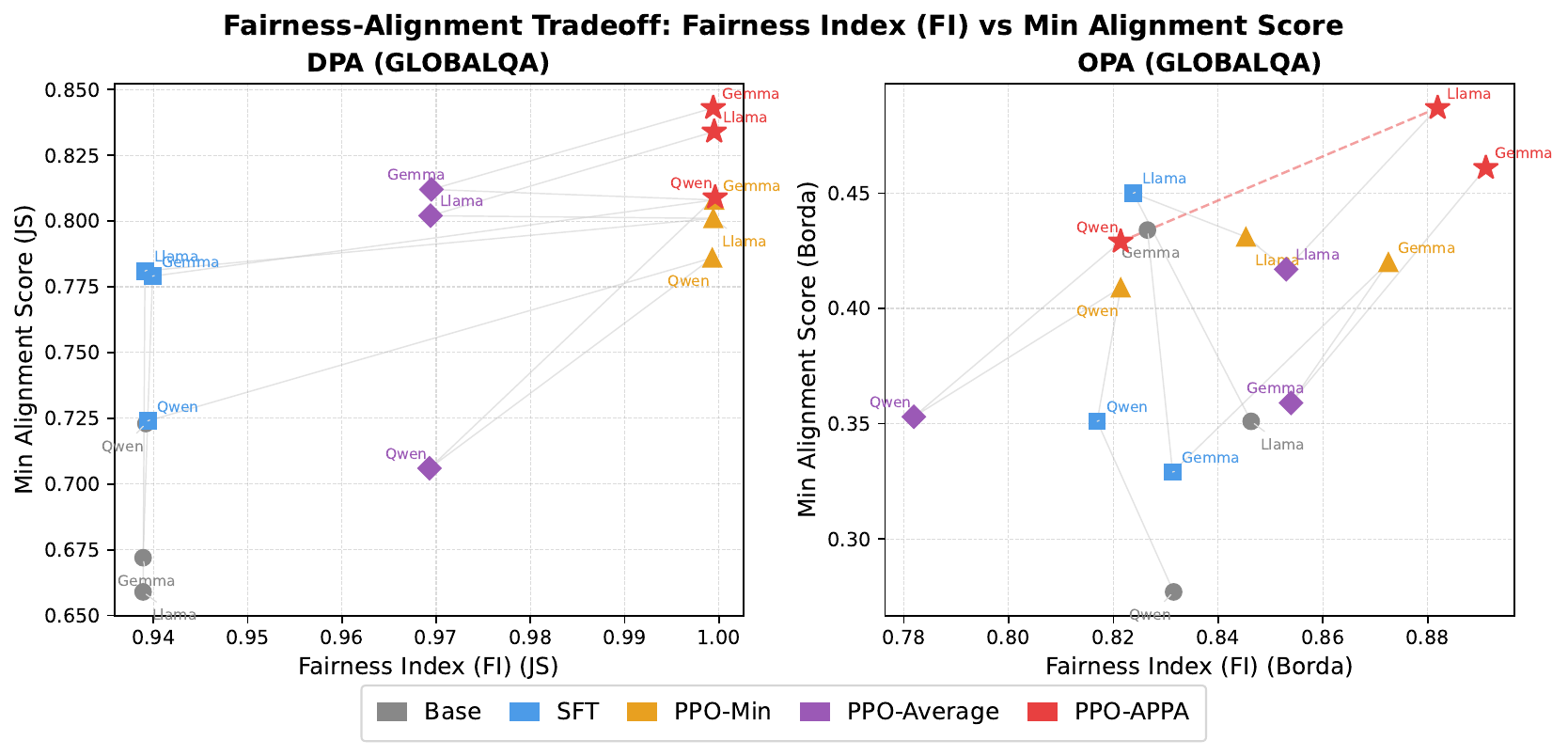}
\caption{%
    \textbf{Fairness--Alignment trade-off: QA Fairness Index vs.\ Minimum
    Alignment Score (GLOBALQA).}
    Left: DPA task using JS metric.
    Right: OPA task using Borda metric.
    Each point is a (model, aggregation strategy) pair.
    PPO-\sysname (red stars) generally occupies the upper-right region
    across the evaluated model families.
    Gray lines connect strategy points within each model family,
    illustrating the progression from Base through SFT to PPO strategies.
}
  \label{fig:tradeoff-scatter}
\end{figure*}


Figure~\ref{fig:tradeoff-scatter} plots the joint $(FI,\text{ Min AS})$ space for all
model--aggregation combinations on GLOBALQA. In the DPA panel, PPO-\sysname{}
occupies the upper-right corner for all three models (FI $\approx 1.00$,
Min AS $\approx 0.81$--$0.84$). Base models cluster at the lower-left
(FI $\approx 0.94$, Min AS $\approx 0.66$--$0.72$), while SFT increases
Min AS modestly with little change in FI. In the OPA panel, PPO-Average
achieves intermediate FI ($\approx 0.78$--$0.86$) but lower Min AS than
PPO-\sysname{} across all models. PPO-Min attains high FI (often matching
or exceeding PPO-Average) but consistently lower Min AS than PPO-\sysname{},
highlighting a trade-off between fairness and worst-group alignment.
The OPA panel shows a similar pattern at lower absolute values, reflecting
greater task difficulty. Gemma-2-2B with PPO-\sysname{} reaches the highest
Borda FI ($0.891$) of any model--strategy pair, with Min Borda AS $0.46$.
PPO-Min typically achieves high FI but lower Min AS, and in the broader
results table this is also accompanied by lower Avg AS. The corresponding
OQA trade-off plot is in Appendix~\ref{appendix:oqa-tradeoff},
Figure~\ref{fig:tradeoff-scatter-oqa}.

\section{Conclusion \& Future Work}
\label{sec:conclusion}

Reward aggregation plays a key role in \acf{fedrlhf}, as it determines how different groups influence the final model. 
Simple averaging tends to favor majority groups, while minimax focuses only on the worst-performing group and ignores useful information from others. \sysname uses adaptive weights based on historical alignment rewards together with a threshold on the Fairness Index ($FI$) to better balance group contributions. It improves worst-group performance by up to $28\%$ over average aggregation while in most cases maintaining higher overall alignment than min aggregation across two datasets, two tasks, and three model families. These results show that how we aggregate rewards is an important design choice in pluralistic \acs{fedrlhf}. Future work includes applying \sysname{} beyond multiple-choice question  answering to tasks such as long form text generation, code synthesis, and  creative writing, where group preferences are richer and less structured, and where the design of group-level reward signals itself remains an open  challenge.

\section*{Acknowledgments}
This work is supported by the U.S. National Science Foundation (NSF) under grant number 2339266.

\bibliographystyle{colm2026_conference}
\bibliography{reference}

\appendix

\section{Extended Related Work}
\label{app:related}

\paragraph{Steering \ac{llms} Toward Group Preferences.} Prompt engineering approaches, including metadata appending, persona prompts, and
in-context few-shot examples, steer \ac{llms} toward group preferences without
updating model weights~\cite{intro3-prompt-no-3, globalopnion}.
While computationally cheap, these methods are heuristic and do not transfer
reliably across model families, and they show only limited gains on challenging
opinion surveys~\cite{gpo, globalopnion}.
\ac{gpo}~\cite{gpo} demonstrated that a lightweight transformer module trained
via meta-learning on few-shot group examples can better capture group preference
distributions than prompt-only approaches.
PluralLLM~\cite{srewa2025pluralllm} extends this idea to a federated setting,
allowing groups to train local preference predictors without sharing raw data.
Both approaches highlight the importance of gradient-based adaptation for
group-level alignment, but neither integrates these predictors into a full
\ac{rlhf} loop.

\paragraph{\ac{ppo}-Based Alignment and Group Robustness.}
\ac{rlhf} with \ac{ppo} has become a standard approach for aligning \ac{llms}
with human preferences~\cite{schulman2017ppo, ouyang2022training}.
In this setup, the policy is updated using rewards from a learned reward model
trained on pairwise comparisons.
While effective for a single population, standard \ac{rlhf} aggregates preferences
into a single reward, which can fail to capture differences across groups~\cite{maxmin}.
MaxMin-RLHF~\cite{maxmin} addresses this by learning multiple reward models,
one per group, and updating the policy using the lowest-performing group at
each step.
This approach focuses entirely on the worst-performing group, discarding
contributions from other groups. ~\cite{bai2022training} similarly introduce weighted group losses for
reward model training, but their method also remains centralized and relies on
reward models.

\paragraph{Reward-Free Alignment and Group Robustness.}
\ac{dpo}~\cite{rafailov2023direct} avoids explicit reward model training by
reformulating the \ac{rlhf} objective.
\ac{grpo}~\cite{ramesh2024group} extends this idea to a group-robust setting by
optimizing a worst-group objective and updating group weights based on per-group loss.
IS-DPO, introduced in the same work, adjusts for group size imbalance through
importance sampling but does not address differences in preference distributions.
These methods improve worst-group performance but remain centralized, requiring
all group data at a single location, and operate on pairwise preferences rather
than group-level probability distributions.

\paragraph{Federated Reward Learning and Our Work.}
Recent work in \ac{fedrlhf} explores ways to keep data local.
FedBiOT~\cite{llmfed} allows clients to fine-tune local adapters on private data,
while FedRLHF~\cite{llmfed-2} enables each client to run a local \ac{rlhf} loop
and share updates with the server.
Both approaches require substantial local training at each round. PluralLLM~\cite{srewa2025pluralllm} follows a different approach: each group trains a lightweight preference predictor once using FedAvg, which then outputs a probability distribution over answer options for each input question at inference time. These outputs serve as group-level rewards for \ac{rlhf} without additional local training. Our work builds on this idea by using PluralLLM group-level rewards within a \ac{ppo}-based \ac{fedrlhf} loop and focusing on how to aggregate these rewards
fairly at the server. Unlike MaxMin-RLHF~\cite{maxmin}, which focuses only on the worst-performing
group at each step and discards contributions from other groups, our approach
assigns non-zero weights to every group at every iteration, improving both
average and worst-group alignment simultaneously.
\section{PluralLLM Preference Probability Modeling}
\label{app:pluralllm}

Our \ac{fedrlhf} setup requires group-specific preference distributions over answer options, rather than scalar rewards, for each question–answer pair. In practice, we do not have complete empirical distributions for every group on every question. For example, in \textsc{GLOBALQA}, each country only answers a subset of questions, so many (group, question) pairs are missing. Similar sparsity appears in other survey datasets, especially for smaller or evaluation-only groups. To address this, we use PluralLLM~\citep{srewa2025pluralllm}, a federated transformer-based model that estimates
\[
p(\text{option} \mid q, g)
\]
for each group $g$ in a privacy-preserving way. PluralLLM is trained once using \ac{fl}, where each training group keeps its local dataset $D_g$ and only shares model updates, which are aggregated using FedAvg. No raw preference data is centralized. After training, a single global PluralLLM model is used in a frozen manner.

At inference time, for each question $q_j$ and group $g$, we sample a small number of examples from the local dataset $D_g$ and use them as few-shot context. Given this context and the question $q_j$, PluralLLM outputs a probability distribution
\[
\hat{p}^{\text{PluralLLM}}_{g,j} \in [0,1]^K
\]
over the $K$ answer options, which we use as the group’s preference distribution. The policy model outputs its own distribution $y^{t,\text{policy}}_j$ over the same options. We then compute group-level rewards $r^t_{g,j}$ by measuring the similarity between $y^{t,\text{policy}}_j$ and $\hat{p}^{\text{PluralLLM}}_{g,j}$ using the reward metrics defined in Appendix~\ref{appendix:metrics}. These rewards are used for aggregation during \ac{ppo} training and for evaluation.

In our experiments, we use two separate PluralLLM models: one trained on \textsc{GLOBALQA} data and another trained on \textsc{OQA} data. Each model is used with its corresponding dataset to produce group-conditioned preference distributions. This setup allows us to obtain consistent preference estimates under realistic data sparsity and privacy constraints, without centralizing raw preference data.

\section{Complete PPO Objective and Reward Integration}
\label{app:ppo}
PPO~\cite{schulman2017ppo} optimizes a composite loss:
\begin{equation}
\label{eq:ppo-full}
L_{\text{PPO}} = L_{\text{POLICY}} + c_1\,L_{\text{VF}} + c_2\,L_{\text{ENTROPY}},
\end{equation}
where
\begin{align}
L_{\text{POLICY}} &= \mathbb{E}_t\!\left[\min\!\left(
    \rho_t\,\hat{A}_t,\;
    \mathrm{clip}\!\left(\rho_t,1-\epsilon,1+\epsilon\right)\hat{A}_t
  \right)\right], \quad
\rho_t = \frac{\pi_\theta(a_t|s_t)}{\pi_{\theta_{\mathrm{old}}}(a_t|s_t)},
\label{eq:ppo-policy}\\
L_{\text{VF}} &= \tfrac{1}{2}\bigl|V_\theta(s)
  - \textstyle\sum_{t=0}^{T}\gamma^t r_t\bigr|^2,
\label{eq:ppo-vf}\\
L_{\text{ENTROPY}} &= -\textstyle\sum_x p(x)\log p(x).
\label{eq:ppo-entropy}
\end{align}
Advantages are estimated via GAE~\cite{schulman2015gae}:
\begin{equation}
\label{eq:gae}
\hat{A}_t = \delta_t + \gamma\lambda_{GAE}\,\hat{A}_{t+1}, \qquad
\delta_t = r_t + \gamma V^\pi(s_{t+1}) - V^\pi(s_t).
\end{equation}

\paragraph{Where the reward lives.}
The reward $r_t$ enters \emph{exclusively} through $\delta_t$
in Equation~\eqref{eq:gae}, propagating into $\hat{A}_t$ and
thence into $L_{\text{POLICY}}$.

\paragraph{Standard RLHF objective.}
Standard RLHF fine-tunes a policy $\pi_\theta$ to maximize a single centralized reward
$r(x,y)$ subject to a KL penalty~\cite{schulman2017ppo,ouyang2022training}:
\begin{equation}
\label{eq:ppo-standard}
\max_{\pi_\theta}\;
\mathbb{E}_{x \sim \mathcal{D},\, y \sim \pi_\theta(\cdot|x)}
\!\left[r(x,y)\;-\;\beta\,
  \mathrm{KL}\!\left[\pi_\theta(\cdot|x)\,\|\,\pi_{\text{ref}}(\cdot|x)\right]
\right].
\end{equation}
In practice, $r_t$ absorbs this KL penalty against a frozen reference policy
to prevent reward hacking~\cite{ouyang2022training}:
\begin{equation}
\label{eq:rlhf-reward}
r(s_t,a_t) \;\leftarrow\; r(s_t,a_t)
  - \beta\,\mathrm{KL}\!\left[\pi_\theta(\cdot|s_t)\,\|\,\pi_{\mathrm{ref}}(\cdot|s_t)\right].
\end{equation}

\paragraph{Pluralistic extension.}
Ours replaces the single centralized reward in
Equation~\eqref{eq:rlhf-reward} with the adaptive aggregated reward
$\mathrm{Agg}_{\boldsymbol{\alpha}^t}(\mathbf{r}^t_j)$ from Equation~\eqref{eq:ppo-federated}.
All other PPO components remain \emph{unchanged}; only $r_t$ in
Equation~\eqref{eq:gae} is substituted, making our federated reward
aggregation a drop-in extension of the standard PPO pipeline.

\section{Dataset Details}
\label{appendix:datasets}

We evaluate on two complementary survey corpora.

\begin{itemize}
  \item \textbf{GLOBALQA} (Pew Research Global Attitudes)~\cite{globalopnion}:
  2,554 multiple-choice questions spanning politics, media, technology,
  religion, race, and ethnicity, with respondents from diverse countries.
  Each country is treated as a separate federated client.
  Questions are predominantly \emph{non-ordinal} (nominal answer sets), so we use Jensen-Shannon divergence as the primary similarity metric for \ac{dpa}.
  For \ac{opa}, we evaluate using Borda score as the canonical metric since
  Borda scoring is appropriate for nominal-ordered rankings.

  \item \textbf{OQA} (OpinionQA)~\cite{intro3-prompt-no-3}:
  A US-centric survey with respondents stratified by demographic group
  (e.g., political affiliation, religion, education, income, race, region).
  Answers are predominantly \emph{ordinal} (e.g., Likert-scale responses),
  making Wasserstein distance the preferred \ac{dpa} metric, since it respects
  the natural ordering of answer options.
  For \ac{opa}, Borda score again serves as the primary evaluation metric.
\end{itemize}

These two datasets offer complementary evaluation regimes:
GLOBALQA tests \emph{cross-national} pluralistic alignment with large group
diversity and nominal answer spaces, while OQA tests \emph{intra-national}
demographic alignment with ordinal preference structure.

\section{Evaluation Metrics}
\label{appendix:metrics}

\subsection{Reward Metrics -- \ac{dpa} Task}

\textbf{Wasserstein Reward:}
\begin{equation}
r^{t,\text{Was}}_{g,j} = 1 - \frac{W_1(y^{t,\text{policy}}_j,\, \hat{p}^{\text{PluralLLM}}_{g,j})}{K-1}
\end{equation}
$r^{t,\text{Was}}_{g,j} \in [0,1]$, where $1$ indicates a perfect distribution match.
It measures the geometric cost of transporting mass between the predicted and
target distributions, penalizing predictions proportionally to how far
probability mass is displaced. It is the primary metric for OQA, where
ordinal answer structure means distant misplacements should incur larger
penalties than adjacent ones.

\textbf{Cosine Similarity Reward:}
\begin{equation}
r^{t,\text{Cos}}_{g,j} =
  \frac{1}{2}\left(1 +
  \frac{y^{t,\text{policy}}_j \cdot \hat{p}^{\text{PluralLLM}}_{g,j}}
       {\|y^{t,\text{policy}}_j\|\,\|\hat{p}^{\text{PluralLLM}}_{g,j}\|}\right)
\end{equation}
$r^{t,\text{Cos}}_{g,j} \in [0,1]$; higher is better.

It captures directional agreement between predicted and target preference
vectors regardless of absolute scale, revealing whether the model's relative
option preferences match the group's even when the predicted probabilities are not perfectly aligned in scale and is normalized $[0,1]$ to remain consistent with the other
reward metrics. It serves as a magnitude-invariant complement to Wasserstein.

\textbf{Jensen-Shannon Reward:}
\begin{equation}
r^{t,\text{JS}}_{g,j} =
  1 - \mathrm{JSD}\!\left(y^{t,\text{policy}}_j \,\big\|\, \hat{p}^{\text{PluralLLM}}_{g,j}\right)
\end{equation}
$r^{t,\text{JS}}_{g,j} \in [0,1]$; higher is better.
It measures symmetric information-theoretic divergence between two
distributions. Being bounded and symmetric, it is a stable alternative to KL
divergence and is the primary metric for GLOBALQA, where nominal answer
options carry no ordinal structure that would favor geometry-sensitive
measures.

\subsection{Reward Metrics -- \ac{opa} Task}

\textbf{Borda Reward:}
\begin{equation}
r^{t,\text{Bor}}_{g,j}
    = \frac{\sum_{k=1}^{K} (K - k + 1)\,
        \mathbb{I}\!\bigl[
            \text{rank}(y^{t,\text{policy}}_{j})_k
            = \text{rank}(\hat{p}^{\text{PluralLLM}}_{g,j})_k
        \bigr]
      }{K(K+1)/2}
\end{equation}
$r^{t,\text{Bor}}_{g,j} \in [0,1]$; $1$ indicates perfect position-wise agreement.
It assigns position-weighted credit, rewarding correct placement of
higher-ranked options more than lower-ranked ones. This makes it more
informative than binary exact-match and more stable than Kendall $\tau$
when the number of options $K$ is small, as is typical in survey datasets.

\subsection{Group, Average, and Minimum Alignment Scores}
\label{appendix:as-def}

Let $\mathcal{Q}_{\text{test}}$ denote the set of test questions used for evaluation.
For each question $q_j \in \mathcal{Q}_{\text{test}}$ and group $g \in G_{\text{train}}$,
we compute a per-question group-level reward $r^{\text{M}}_{g,j} \in [0,1]$ under metric
$\text{M} \in \{\text{JS}, \text{Was.}, \text{Cos.}, \text{Bor.}\}$ as defined above.

\paragraph{Per-group alignment score (AS).}
The alignment score for a fixed group $g$ is the average reward over all test Q/A pairs:
\begin{equation}
\label{eq:group-as}
AS^{\text{M}}_{g}
\;=\;
\frac{1}{|\mathcal{Q}_{\text{test}}|}
\sum_{q_j \in \mathcal{Q}_{\text{test}}}
r^{\text{M}}_{g,j}.
\end{equation}

\paragraph{Average and minimum alignment scores.}
We summarize performance across groups by taking the mean and minimum over the per-group scores:
\begin{equation}
\label{eq:avg-min-as}
\mathrm{AvgAS}^{\text{M}}
\;=\;
\frac{1}{|G_{\text{train}}|}
\sum_{g \in G_{\text{train}}} AS^{\text{M}}_{g},
\qquad
\mathrm{MinAS}^{\text{M}}
\;=\;
\min_{g \in G_{\text{train}}} AS^{\text{M}}_{g}.
\end{equation}
Thus, the reported \emph{Avg AS} and \emph{Min AS} are computed over per-group averages
(Equation~\eqref{eq:group-as}), not directly over individual questions.

\paragraph{Relationship to aggregation during training.}
The adaptive alpha aggregation $\mathrm{Agg}_{\boldsymbol{\alpha}^t}(\cdot)$ used during PPO training is
defined in the proposed method (Section~\ref{sec:adaptive-alpha}); here we only define how
we compute and summarize per-group alignment scores at evaluation time.
\section{Full Quantitative Results}
\label{appendix:tables}

Tables~\ref{tab:combined-value-ppo-base-sft} and~\ref{tab:combined-order-ppo-base-sft}
report full DPA and OPA results across all models, metrics, and aggregation strategies.

\newcommand{\na}{\textit{N/A}}

\begin{table*}[t]
\centering
\caption{Distributional Preference Alignment (DPA) (GLOBALQA + OQA, Base vs SFT vs PPO). Reward-wise QA fairness and alignment metrics.}
\label{tab:combined-value-ppo-base-sft}
\renewcommand{\arraystretch}{1.2}
\setlength{\tabcolsep}{4pt}
\setlength{\extrarowheight}{1pt}
\small
\resizebox{\textwidth}{!}{%
\begin{tabular}{|>{\centering\arraybackslash}m{1.9cm}|>{\centering\arraybackslash}m{3.1cm}|>{\centering\arraybackslash}m{1.5cm}|>{\centering\arraybackslash}m{1.95cm}|>{\centering\arraybackslash}m{1.05cm}|>{\centering\arraybackslash}m{1.05cm}|>{\centering\arraybackslash}m{1.05cm}|>{\centering\arraybackslash}m{1.05cm}|>{\centering\arraybackslash}m{1.05cm}|>{\centering\arraybackslash}m{1.05cm}|>{\centering\arraybackslash}m{1.05cm}|>{\centering\arraybackslash}m{1.05cm}|>{\centering\arraybackslash}m{1.05cm}|>{\centering\arraybackslash}m{1.15cm}|}
\hline
\multirow{2}{*}{\textbf{Dataset}} & \multirow{2}{*}{\textbf{Model}} & \multirow{2}{*}{\textbf{Method}} & \multirow{2}{*}{\textbf{Server Agg.}} & \multicolumn{3}{|c|}{\textbf{FI}} & \multicolumn{3}{|c|}{\textbf{Avg AS}} & \multicolumn{3}{|c|}{\textbf{Min AS}} & \multicolumn{1}{c|}{\multirow{2}{*}{\shortstack{\textbf{Format}\\\textbf{Score}}}} \\
\hhline{~~~~---------~}
& & & & \textbf{JS$\uparrow$} & \textbf{Was.$\uparrow$} & \textbf{Cos.$\uparrow$} & \textbf{JS$\uparrow$} & \textbf{Was.$\uparrow$} & \textbf{Cos.$\uparrow$} & \textbf{JS$\uparrow$} & \textbf{Was.$\uparrow$} & \textbf{Cos.$\uparrow$} \\
\hline
\multirow{15}{*}{GLOBALQA} & \multirow{5}{*}{gemma-2-2b} & BASE & - & 0.9389 & 0.9392 & 0.9368 & 0.682 & 0.783 & 0.754 & 0.672 & 0.773 & 0.739 & 0.968 \\
\cline{3-14}
& & SFT & - & 0.9399 & 0.9394 & 0.9389 & 0.786 & 0.847 & 0.855 & 0.779 & 0.841 & 0.840 & 0.958 \\
\cline{3-14}
& & \multirow{3}{*}{PPO/JS} & MIN & \textbf{0.9995} & 0.9997 & \textbf{0.9996} & 0.834 & 0.886 & 0.909 & 0.808 & 0.862 & 0.885 & \textbf{0.972} \\
\cline{4-4}\cline{5-14}
& &  & AVERAGE & 0.9695 & 0.9697 & 0.9696 & 0.839 & 0.890 & 0.914 & 0.812 & 0.868 & 0.890 & 0.957 \\
\cline{4-4}\cline{5-14}
& &  & \sysname (OURS) & 0.9994 & \textbf{0.9997} & 0.9995 & \textbf{0.861} & \textbf{0.912} & \textbf{0.916} & \textbf{0.843} & \textbf{0.898} & \textbf{0.891} & 0.963 \\
\noalign{\global\setlength{\arrayrulewidth}{1.3pt}}
\cline{2-14}
\noalign{\global\setlength{\arrayrulewidth}{0.4pt}}
 & \multirow{5}{*}{llama-3.2-3b} & BASE & - & 0.9389 & 0.9390 & 0.9345 & 0.664 & 0.745 & 0.710 & 0.659 & 0.741 & 0.696 & 0.958 \\
\cline{3-14}
& & SFT & - & 0.9391 & 0.9393 & 0.9381 & 0.785 & 0.845 & 0.862 & 0.781 & 0.844 & 0.857 & \textbf{0.981} \\
\cline{3-14}
& & \multirow{3}{*}{PPO/JS} & MIN & 0.9994 & 0.9996 & \textbf{0.9996} & 0.826 & 0.876 & 0.908 & 0.801 & 0.854 & \textbf{0.886} & 0.962 \\
\cline{4-4}\cline{5-14}
& &  & AVERAGE & 0.9694 & 0.9696 & 0.9695 & 0.826 & 0.879 & 0.905 & 0.802 & 0.857 & 0.883 & 0.962 \\
\cline{4-4}\cline{5-14}
& &  & \sysname (OURS) & \textbf{0.9995} & \textbf{0.9996} & 0.9995 & \textbf{0.848} & \textbf{0.899} & \textbf{0.908} & \textbf{0.834} & \textbf{0.888} & 0.885 & 0.977 \\
\noalign{\global\setlength{\arrayrulewidth}{1.3pt}}
\cline{2-14}
\noalign{\global\setlength{\arrayrulewidth}{0.4pt}}
 & \multirow{5}{*}{Qwen3-0.6B-Instruct} & BASE & - & 0.9392 & 0.9392 & 0.9369 & 0.730 & 0.791 & 0.762 & 0.723 & 0.788 & 0.750 & \textbf{0.940} \\
\cline{3-14}
& & SFT & - & 0.9394 & 0.9392 & 0.9369 & 0.730 & 0.791 & 0.763 & 0.724 & 0.788 & 0.750 & 0.933 \\
\cline{3-14}
& & \multirow{3}{*}{PPO/JS} & MIN & 0.9993 & 0.9996 & 0.9992 & 0.810 & 0.862 & 0.877 & 0.786 & 0.840 & 0.854 & 0.905 \\
\cline{4-4}\cline{5-14}
& &  & AVERAGE & 0.9693 & 0.9694 & 0.9671 & 0.726 & 0.785 & 0.759 & 0.706 & 0.762 & 0.727 & 0.889 \\
\cline{4-4}\cline{5-14}
& &  & \sysname (OURS) & \textbf{0.9996} & \textbf{0.9996} & \textbf{0.9993} & \textbf{0.824} & \textbf{0.884} & \textbf{0.882} & \textbf{0.809} & \textbf{0.871} & \textbf{0.857} & 0.896 \\
\noalign{\global\setlength{\arrayrulewidth}{2.3pt}}
\hline
\noalign{\global\setlength{\arrayrulewidth}{0.4pt}}
\multirow{15}{*}{OQA} & \multirow{5}{*}{gemma-2-2b} & BASE & - & 0.9285 & 0.9308 & 0.9016 & 0.650 & 0.763 & 0.679 & 0.531 & 0.658 & 0.551 & 0.990 \\
\cline{3-14}
& & SFT & - & 0.9318 & 0.9337 & 0.9288 & 0.768 & 0.835 & 0.835 & 0.620 & 0.714 & 0.700 & 0.979 \\
\cline{3-14}
& & \multirow{3}{*}{PPO/Was.} & MIN & 0.9923 & \textbf{0.9946} & 0.9925 & 0.796 & 0.858 & 0.869 & 0.736 & 0.828 & 0.816 & \textbf{0.991} \\
\cline{4-4}\cline{5-14}
& &  & AVERAGE & 0.9621 & 0.9644 & 0.9632 & 0.797 & 0.860 & \textbf{0.876} & 0.735 & 0.823 & 0.819 & 0.973 \\
\cline{4-4}\cline{5-14}
& &  & \sysname (OURS) & \textbf{0.9924} & 0.9942 & \textbf{0.9939} & \textbf{0.814} & \textbf{0.872} & 0.871 & \textbf{0.766} & \textbf{0.842} & \textbf{0.823} & 0.985 \\
\noalign{\global\setlength{\arrayrulewidth}{1.3pt}}
\cline{2-14}
\noalign{\global\setlength{\arrayrulewidth}{0.4pt}}
 & \multirow{5}{*}{llama-3.2-3b} & BASE & - & 0.9270 & 0.9221 & 0.8886 & 0.614 & 0.681 & 0.629 & 0.499 & 0.561 & 0.477 & 0.989 \\
\cline{3-14}
& & SFT & - & 0.9283 & 0.9247 & 0.8983 & 0.654 & 0.722 & 0.683 & 0.539 & 0.601 & 0.532 & 0.988 \\
\cline{3-14}
& & \multirow{3}{*}{PPO/Was.} & MIN & 0.9918 & 0.9937 & 0.9902 & 0.803 & 0.856 & \textbf{0.878} & 0.732 & 0.811 & 0.792 & \textbf{1.000} \\
\cline{4-4}\cline{5-14}
& &  & AVERAGE & 0.9621 & 0.9645 & 0.9617 & 0.801 & 0.859 & 0.877 & 0.735 & 0.824 & 0.806 & 0.996 \\
\cline{4-4}\cline{5-14}
& &  & \sysname (OURS) & \textbf{0.9922} & \textbf{0.9940} & \textbf{0.9923} & \textbf{0.816} & \textbf{0.872} & 0.872 & \textbf{0.765} & \textbf{0.841} & \textbf{0.815} & \textbf{1.000} \\
\noalign{\global\setlength{\arrayrulewidth}{1.3pt}}
\cline{2-14}
\noalign{\global\setlength{\arrayrulewidth}{0.4pt}}
 & \multirow{5}{*}{Qwen3-0.6B-Instruct} & BASE & - & 0.9309 & 0.9274 & 0.9095 & 0.684 & 0.741 & 0.695 & 0.571 & 0.623 & 0.555 & 0.967 \\
\cline{3-14}
& & SFT & - & 0.9329 & 0.9384 & 0.9105 & 0.704 & 0.781 & 0.699 & 0.601 & 0.653 & 0.583 & 0.967 \\
\cline{3-14}
& & \multirow{3}{*}{PPO/Was.} & MIN & 0.9902 & \textbf{0.9926} & \textbf{0.9828} & \textbf{0.748} & \textbf{0.823} & \textbf{0.812} & 0.685 & \textbf{0.781} & \textbf{0.707} & 0.921 \\
\cline{4-4}\cline{5-14}
& &  & AVERAGE & 0.9610 & 0.9596 & 0.9448 & 0.722 & 0.786 & 0.742 & 0.675 & 0.734 & 0.625 & 0.966 \\
\cline{4-4}\cline{5-14}
& &  & \sysname (OURS) & \textbf{0.9909} & 0.9887 & 0.9717 & 0.722 & 0.780 & 0.720 & \textbf{0.686} & 0.735 & 0.596 & \textbf{1.000} \\
\hline
\end{tabular}
}
\end{table*}

\begin{table*}[t]
\centering
\caption{Ordinal Preference Alignment (OPA) (GLOBALQA + OQA, Base vs SFT vs PPO). Reward-wise QA fairness and alignment metrics.}
\label{tab:combined-order-ppo-base-sft}
\renewcommand{\arraystretch}{1.2}
\setlength{\tabcolsep}{4pt}
\setlength{\extrarowheight}{1pt}
\small
\resizebox{\textwidth}{!}{%
\begin{tabular}{|>{\centering\arraybackslash}m{1.9cm}|>{\centering\arraybackslash}m{3.1cm}|>{\centering\arraybackslash}m{1.5cm}|>{\centering\arraybackslash}m{1.95cm}|>{\centering\arraybackslash}m{1.05cm}|>{\centering\arraybackslash}m{1.05cm}|>{\centering\arraybackslash}m{1.05cm}|>{\centering\arraybackslash}m{1.15cm}|}
\hline
\multirow{2}{*}{\textbf{Dataset}} & \multirow{2}{*}{\textbf{Model}} & \multirow{2}{*}{\textbf{Method}} & \multirow{2}{*}{\textbf{Server Agg.}} & \multicolumn{1}{|c|}{\textbf{FI}} & \multicolumn{1}{|c|}{\textbf{Avg AS}} & \multicolumn{1}{|c|}{\textbf{Min AS}} & \multicolumn{1}{c|}{\multirow{2}{*}{\shortstack{\textbf{Format}\\\textbf{Score}}}} \\
\hhline{~~~~---~}
& & & & \textbf{Bor.} & \textbf{Bor.} & \textbf{Bor.} \\
\hline
\multirow{15}{*}{GLOBALQA} & \multirow{5}{*}{gemma-2-2b} & BASE & - & 0.8265 & 0.461 & 0.434 & 0.999 \\
\cline{3-8}
& & SFT & - & 0.8313 & 0.339 & 0.329 & \textbf{1.000} \\
\cline{3-8}
& & \multirow{3}{*}{PPO/Bor.} & MIN & 0.8725 & 0.475 & 0.420 & \textbf{1.000} \\
\cline{4-4}\cline{5-8}
& &  & AVERAGE & 0.8539 & 0.469 & 0.359 & \textbf{1.000} \\
\cline{4-4}\cline{5-8}
& &  & \sysname (OURS) & \textbf{0.8911} & \textbf{0.511} & \textbf{0.461} & \textbf{1.000} \\
\noalign{\global\setlength{\arrayrulewidth}{1.3pt}}
\cline{2-8}
\noalign{\global\setlength{\arrayrulewidth}{0.4pt}}
 & \multirow{5}{*}{llama-3.2-3b} & BASE & - & 0.8463 & 0.363 & 0.351 & 0.771 \\
\cline{3-8}
& & SFT & - & 0.8238 & 0.461 & 0.450 & 0.934 \\
\cline{3-8}
& & \multirow{3}{*}{PPO/Bor.} & MIN & 0.8453 & 0.451 & 0.431 & \textbf{0.969} \\
\cline{4-4}\cline{5-8}
& &  & AVERAGE & 0.8530 & 0.526 & 0.417 & \textbf{0.969} \\
\cline{4-4}\cline{5-8}
& &  & \sysname (OURS) & \textbf{0.8819} & \textbf{0.536} & \textbf{0.487} & \textbf{0.969} \\
\noalign{\global\setlength{\arrayrulewidth}{1.3pt}}
\cline{2-8}
\noalign{\global\setlength{\arrayrulewidth}{0.4pt}}
 & \multirow{5}{*}{Qwen3-0.6B-Instruct} & BASE & - & \textbf{0.8315} & 0.326 & 0.277 & 0.959 \\
\cline{3-8}
& & SFT & - & 0.8169 & 0.360 & 0.351 & \textbf{1.000} \\
\cline{3-8}
& & \multirow{3}{*}{PPO/Bor.} & MIN & 0.8214 & 0.422 & 0.409 & \textbf{1.000} \\
\cline{4-4}\cline{5-8}
& &  & AVERAGE & 0.7819 & 0.416 & 0.353 & \textbf{1.000} \\
\cline{4-4}\cline{5-8}
& &  & \sysname (OURS) & 0.8214 & \textbf{0.442} & \textbf{0.429} & \textbf{1.000} \\
\noalign{\global\setlength{\arrayrulewidth}{2.3pt}}
\hline
\noalign{\global\setlength{\arrayrulewidth}{0.4pt}}
\multirow{15}{*}{OQA} & \multirow{5}{*}{gemma-2-2b} & BASE & - & 0.5898 & 0.383 & 0.255 & \textbf{1.000} \\
\cline{3-8}
& & SFT & - & 0.6311 & 0.471 & 0.318 & \textbf{1.000} \\
\cline{3-8}
& & \multirow{3}{*}{PPO/Bor.} & MIN & 0.7067 & 0.500 & 0.312 & \textbf{1.000} \\
\cline{4-4}\cline{5-8}
& &  & AVERAGE & 0.6787 & \textbf{0.518} & 0.309 & \textbf{1.000} \\
\cline{4-4}\cline{5-8}
& &  & \sysname (OURS) & \textbf{0.7141} & 0.503 & \textbf{0.315} & \textbf{1.000} \\
\noalign{\global\setlength{\arrayrulewidth}{1.3pt}}
\cline{2-8}
\noalign{\global\setlength{\arrayrulewidth}{0.4pt}}
 & \multirow{5}{*}{llama-3.2-3b} & BASE & - & 0.6292 & 0.364 & 0.231 & 0.910 \\
\cline{3-8}
& & SFT & - & 0.6284 & 0.402 & 0.253 & 0.958 \\
\cline{3-8}
& & \multirow{3}{*}{PPO/Bor.} & MIN & 0.6786 & 0.434 & \textbf{0.294} & \textbf{1.000} \\
\cline{4-4}\cline{5-8}
& &  & AVERAGE & 0.6499 & 0.475 & 0.270 & \textbf{1.000} \\
\cline{4-4}\cline{5-8}
& &  & \sysname (OURS) & \textbf{0.7213} & \textbf{0.499} & 0.285 & \textbf{1.000} \\
\noalign{\global\setlength{\arrayrulewidth}{1.3pt}}
\cline{2-8}
\noalign{\global\setlength{\arrayrulewidth}{0.4pt}}
 & \multirow{5}{*}{Qwen3-0.6B-Instruct} & BASE & - & 0.5543 & 0.334 & 0.218 & 0.993 \\
\cline{3-8}
& & SFT & - & 0.5906 & 0.366 & \textbf{0.239} & \textbf{1.000} \\
\cline{3-8}
& & \multirow{3}{*}{PPO/Bor.} & MIN & 0.6544 & 0.428 & 0.237 & \textbf{1.000} \\
\cline{4-4}\cline{5-8}
& &  & AVERAGE & 0.6215 & 0.424 & 0.210 & \textbf{1.000} \\
\cline{4-4}\cline{5-8}
& &  & \sysname (OURS) & \textbf{0.7037} & \textbf{0.450} & 0.230 & \textbf{1.000} \\
\hline
\end{tabular}
}
\end{table*}

\begin{figure}[t]
\centering
\fbox{%
\begin{minipage}{0.88\linewidth}
\small
\textbf{Distributional Preference Alignment Prompt}

\noindent\rule{\linewidth}{0.4pt}

\noindent\textit{You are an expert in modelling group preferences. You will
receive a question and $K$ answer options, where $K$ varies per question.}

\medskip
\noindent\textbf{Task:}
\begin{itemize}[leftmargin=1.2em, topsep=2pt, itemsep=1pt]
  \item Assign a preference score to each and every option.
  \item Produce exactly $K$ scores — no option may be skipped or combined.
  \item Each score must be a decimal in $[0,1]$, and the rounded scores must sum to 1.00.
  \item Higher scores represent options a typical group is more likely to choose.
\end{itemize}

\medskip
\noindent\textbf{Output format:}
\begin{itemize}[leftmargin=1.2em, topsep=2pt, itemsep=1pt]
  \item One line, comma-separated decimals, no spaces.
  \item Round each value to 2 decimal places.
  \item No extra text, labels, or symbols.
  \item Example ($K$=4): \texttt{0.65,0.20,0.10,0.05}
\end{itemize}

\noindent\rule{\linewidth}{0.4pt}

\noindent\textbf{Question:} Germany's influence in the EU\\[3pt]
\noindent\textbf{Options:}\\
\texttt{A:} Has too much influence \quad
\texttt{B:} Has too little influence \\
\texttt{C:} Has about the right amount \quad
\texttt{D:} DK/Refused

\medskip
\noindent Return \textbf{only} the $K$ scores in the same order as the options.
\end{minipage}%
}
\caption{Distributional Preference Alignment (DPA) prompt template. The model
outputs a calibrated probability distribution over all answer options. The
number of options $K$ varies dynamically per question.}
\label{fig:prob-prompt}
\end{figure}

\begin{figure}[t]
\centering
\fbox{%
\begin{minipage}{0.88\linewidth}
\small
\textbf{Ordinal Preference Alignment Prompt}

\noindent\rule{\linewidth}{0.4pt}

\noindent\textit{You are an expert in ranking group preferences. You will
receive a question and $K$ answer options, where $K$ varies per question.}

\medskip
\noindent\textbf{Task:}
\begin{itemize}[leftmargin=1.2em, topsep=2pt, itemsep=1pt]
  \item Rank all $K$ options from most to least preferred.
  \item Process every option — no skipping or combining.
  \item Order options based on what a typical group would most likely choose.
  \item Most preferred option appears first.
\end{itemize}

\medskip
\noindent\textbf{Output format:}
\begin{itemize}[leftmargin=1.2em, topsep=2pt, itemsep=1pt]
  \item One line, comma-separated option letters, no spaces.
  \item Use the exact letters provided in the question.
  \item No extra text, labels, or symbols.
  \item Example ($K$=4): \texttt{B,C,A,D}
\end{itemize}

\noindent\rule{\linewidth}{0.4pt}

\noindent\textbf{Question:} Germany's influence in the EU\\[3pt]
\noindent\textbf{Options:}\\
\texttt{A:} Has too much influence \quad
\texttt{B:} Has too little influence \\
\texttt{C:} Has about the right amount \quad
\texttt{D:} DK/Refused

\medskip
\noindent Return \textbf{only} the $K$-letter ranking from most to least preferred.
\end{minipage}%
}
\caption{Ordinal Preference Alignment (OPA) prompt template. The model outputs
a ranked ordering of all answer options from most to least preferred. The
number of options $K$ varies dynamically per question.}
\label{fig:rank-prompt}
\end{figure}
\definecolor{modelgray}{gray}{0.90}
\definecolor{alphawins}{RGB}{214, 234, 214}

\section{Experiment Configurations and Hyperparameters}
\label{appendix:setup}

Our experimental setup begins with supervised fine-tuning (SFT) as outlined in Table~\ref{tab:sft-hparams}.
We apply LoRA adaptation with rank 16 across all model families for efficient parameter updates.
SFT training uses a cosine learning rate scheduler with warmup ratio $0.1$ and is conducted
for one epoch (GLOBALQA) or four epochs (OQA) to establish our baseline models.

\begin{table}[h]
\centering
\caption{SFT configuration and hyperparameters.}
\label{tab:sft-hparams}
\begin{tabular}{l l}
\toprule
\textbf{Hyperparameter} & \textbf{Value} \\
\midrule
\multicolumn{2}{l}{\textit{Models}} \\
Base models & \texttt{gemma-2-2b-it}, \texttt{Llama-3.2-3 B-Instruct}, \\ &  \texttt{Qwen3-0.6B-Instruct} \\
Precision & BF16 \\
\addlinespace[2pt]
\multicolumn{2}{l}{\textit{Data / Task}} \\
Train/valid split & 80/20\% \\
Max sequence length & 500 tokens (prompt + response) \\
Loss masking & Response tokens only (query tokens masked) \\
\addlinespace[2pt]
\multicolumn{2}{l}{\textit{LoRA Adapter}} \\
Rank ($r$) & 16 \\
Alpha & 32 \\
Dropout & 0.05 \\
Target modules & \texttt{q,k,v,o\_proj}, \texttt{gate,up,down\_proj} \\
\addlinespace[2pt]
\multicolumn{2}{l}{\textit{Optimization}} \\
Batch size (per device) & 8 \\
Gradient accumulation steps & 2 \\
Effective batch size & 16 \\
Learning rate & $5\times10^{-6}$ \\
Scheduler & cosine \\
Warmup ratio & 0.1 \\
Weight decay & 0.01 \\
\addlinespace[2pt]
\multicolumn{2}{l}{\textit{Training}} \\
Epochs & 1 (GLOBALQA) \;/\; 4 (OQA) \\
\bottomrule
\end{tabular}
\end{table}

\begin{table}[h]
\centering
\caption{PPO configuration and hyperparameters (policy and value models initialized from the respective SFT checkpoints).}
\label{tab:ppo-hparams}
\begin{tabular}{l l}
\toprule
\textbf{Hyperparameter} & \textbf{Value} \\
\midrule
\multicolumn{2}{l}{\textit{General}} \\
Policy model  & \texttt{gemma-2-2b-it} (SFT), \texttt{Llama-3.2-3B-Instruct} (SFT),\\ & \texttt{Qwen3-0.6B-Instruct} (SFT) \\
Value model   & \texttt{gemma-2-2b-it} (SFT), \texttt{Llama-3.2-3B-Instruct} (SFT), \\ &\texttt{Qwen3-0.6B-Instruct} (SFT) \\
\addlinespace[2pt]
\multicolumn{2}{l}{\textit{Model / Quantization}} \\
Quantization & 4-bit (\texttt{nf4}, double-quant = True) \\
Compute dtype & BF16 \\
Attention implementation & eager \\
\addlinespace[2pt]
\multicolumn{2}{l}{\textit{LoRA (PEFT — policy \& value)}} \\
Rank ($r$) & 16 \\
Alpha & 32 \\
Dropout & 0.05 \\
Policy target modules & \texttt{q,k,v,o\_proj} \\
Value target modules & \texttt{q,k,v,o\_proj} + \texttt{score} \\
\addlinespace[2pt]
\multicolumn{2}{l}{\textit{Optimization}} \\
Per-device train batch size & 4 \\
Gradient accumulation steps & 24 \\
Effective batch size & 96 \\
Learning rate & $1\times10^{-5}$ \\
Optimizer & AdamW \\
Weight decay & 0.0 \\
Scheduler & linear \\
Training epochs & 5 \\
\addlinespace[2pt]
\multicolumn{2}{l}{\textit{PPO Trainer}} \\
PPO epochs & 2 \\
Mini-batches & 8 \\
Per-device eval batch size & 64 \\
Response length & 60 tokens \\
Temperature & 0.6 \\
KL coefficient & 0.05 \\
Clip range & 0.2 \\
Clip range (value) & 0.2 \\
Value loss coefficient ($v_f$) & 0.2 \\
Discount factor ($\gamma$) & 1.0 \\
GAE lambda ($\lambda_{GAE}$) & 0.95 \\
Reward whitening & Per rollout (before PPO update)\footnotemark \\
\addlinespace[2pt]
\bottomrule
\end{tabular}
\end{table}
\footnotetext{Rewards are whitened over each rollout batch before the PPO update step.}

\begin{table}[h]
\centering
\caption{\sysname-specific hyperparameters, fixed across all experiments.}
\label{tab:appa-hparams}
\begin{tabular}{l l}
\toprule
\textbf{Hyperparameter} & \textbf{Value} \\
\midrule
EMA decay ($\lambda$) & $0.8$ \\
Softmax temperature ($T$) & $0.1$ \\
Fairness Index threshold ($\tau$) & $0.99$ \\
Initial historical alignment reward ($h_g^0$)& $0$ (all groups) \\
\addlinespace[2pt]
Alignment weight ($\omega$) & $0.85$ \\
\addlinespace[2pt]
Min.\ mean threshold ($\mu_{\min}$) & $1\times10^{-6}$ \\
Max.\ CoV cap ($\mathrm{CoV}_{\max}$) & $10.0$ \\
\bottomrule
\end{tabular}
\end{table}
We fix $\tau{=}0.99$, $\lambda{=}0.8$, and $T{=}0.1$ across all experiments;
a more exhaustive exploration of these hyperparameters is left for future work.
The threshold $\tau{=}0.99$ was selected to ensure the adaptive log-sum-exp
aggregation remains active throughout the majority of training: empirically,
terminal FI values under \sysname converge to $\geq 0.999$ on GLOBALQA DPA,
meaning the aggregation switches to a plain average only once alignment has
effectively converged across all groups. A lower threshold would trigger the
plain-average fallback prematurely, undermining the adaptive mechanism early
in training; a threshold of $1.0$ would never trigger the fallback, risking unnecessary adjustments once inter-group disparity becomes negligible.

\paragraph{Numerical Safeguards in Fairness Index Computation.}
\label{appendix:fi-safeguards}

The CoV in Equation~\eqref{eq:fi} is undefined when the mean reward
$\mu(\{r^t_{g,j}\})$ is zero or near zero, which can occur early in training
when all groups receive identically low rewards for a given question.
We handle this with two safeguards. First, if the standard deviation across
group rewards is exactly zero (all groups receive identical rewards for a
question), we set $\mathrm{CoV}=0$ and $FI=1.0$ directly, reflecting that
identical rewards imply maximum fairness. Second, if the mean reward falls
below a minimum threshold $\mu_{\min}$, the question is excluded from the
FI average for that iteration, preventing CoV explosion from a near-zero
denominator. Additionally, CoV values are capped at a maximum $\mathrm{CoV}_{\max}$
to guard against pathological outliers. These safeguards affect only a small
fraction of questions early in training and do not alter the qualitative
behavior of the adaptive aggregation.

As summarized in Tables~\ref{tab:sft-hparams} and~\ref{tab:ppo-hparams}, both the policy and
value models in PPO are initialized from the respective SFT checkpoint for each model family.
During training, we employ two distinct prompt formats: a \ac{dpa} requiring a probability distribution over all options and an \ac{opa} requiring a complete ranking from most to least preferred
(see Figures~\ref{fig:prob-prompt} and~\ref{fig:rank-prompt}).
In both cases, the cross-entropy loss is computed over response tokens only;
query tokens are masked and contribute no gradient signal, ensuring the model
is trained to predict the target output format (probability vector or ranked ordering) rather than to reproduce the input prompt.
Our implementation builds on the Hugging Face TRL library~\cite{vonwerra2022trl}.
All experiments were conducted on 3 nodes, each equipped with an NVIDIA A100 
GPU, Intel Xeon Gold 6326 CPUs (2.90 GHz), and 256 GB RAM. Although our 
cluster comprises three A100 nodes, each individual experiment (i.e., each 
model--dataset--aggregation combination) is run on a single node with one A100 
GPU; the three nodes are used to run experiments in parallel, not for 
distributed training within a single run.
\section{Additional Fairness-Alignment Trade-off Plots}
\label{appendix:oqa-tradeoff}

\begin{figure*}[t]
  \centering
  \includegraphics[width=\textwidth]{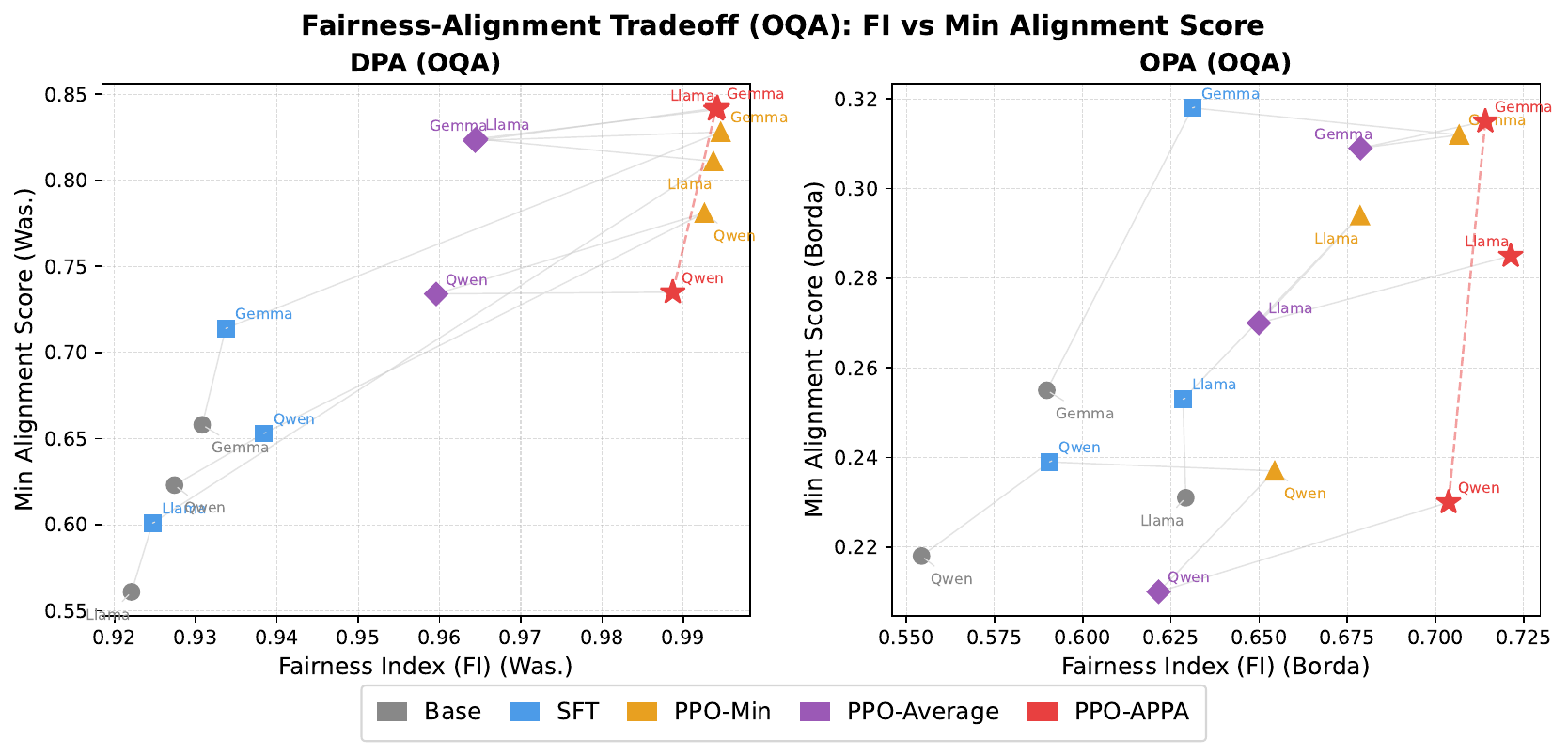}
  \caption{%
    \textbf{Fairness-Alignment Trade-off: QA Fairness Index vs.\ Minimum
    Alignment Score (OQA).}
    Left: DPA task using Wasserstein distance.
    Right: OPA task using Borda metric.
    Each point is a (model, aggregation strategy) pair.
    PPO-\sysname{} (red stars) generally occupies the upper-right, near-Pareto-optimal
    region across the evaluated model families on OQA as well.
    Gray lines connect strategy points within each model family,
    illustrating the progression from Base through SFT to PPO strategies.
  }
  \label{fig:tradeoff-scatter-oqa}
\end{figure*}

Figure~\ref{fig:tradeoff-scatter-oqa} plots the fairness-alignment trade-off on OQA,
showing joint ($FI$, Min AS) space for all model--aggregation combinations.
In the DPA panel (Wasserstein), \textsc{\sysname} again lies toward the upper-right for
Gemma-2-2B and Llama-3.2-3B, combining high FI with the highest minimum alignment
scores across demographic groups, while Base and SFT cluster at lower FI and substantially
lower Min AS.
PPO-Average occupies an intermediate region with improved Min AS over SFT but lower FI
than \textsc{\sysname}.
PPO-Min achieves high FI but lower average alignment than \textsc{\sysname} for
Gemma-2-2B and Llama-3.2-3B; for Qwen3-0.6B, PPO-Min leads on both FI ($0.993$) and
Avg AS ($0.823$), consistent with the exception noted in
Section~\ref{subsubsec:dpa-oqa}.

The OPA panel shows the same qualitative pattern at lower absolute values, reflecting the
greater difficulty of ranking.
\textsc{\sysname} attains the highest Borda FI across all three models
($0.714$ / $0.721$ / $0.704$ for Gemma-2-2B / Llama-3.2-3B / Qwen3-0.6B) and the best
joint (FI, Min AS) for Gemma-2-2B.
For Llama-3.2-3B and Qwen3-0.6B, PPO-Min achieves higher Min AS ($0.294$ and $0.237$
vs.\ $0.285$ and $0.230$) but at the cost of substantially lower FI ($0.679$ and $0.654$
vs.\ $0.721$ and $0.704$), illustrating that \textsc{\sysname} better balances the
fairness--alignment trade-off at the mean level.
PPO-Average fails to dominate on either axis relative to \textsc{\sysname}.

\section{Model Descriptions}
\label{appendix:models}

We evaluate \sysname across three instruction-tuned language models ranging
from 0.6B to 3B parameters, deliberately spanning a wide capacity range to
test whether adaptive reward aggregation generalises across model scales.

\paragraph{Gemma-2-2B-Instruct~\cite{gemma2}.}
Gemma-2-2B is a 2-billion-parameter model from Google DeepMind's Gemma 2
family, trained with knowledge distillation from larger models in the same
family. Despite its compact size, Gemma-2-2B achieves strong performance on
reasoning and instruction-following benchmarks relative to models of
comparable scale. We use the instruction-tuned variant
(\texttt{gemma-2-2b-it}) as our base model for SFT and PPO.

\paragraph{Llama-3.2-3B~\cite{llama3}.}
Llama-3.2-3B is a 3-billion-parameter model from Meta's Llama 3.2 series,
optimized for multilingual dialogue and instruction-following. It represents
the largest model in our evaluation and serves as the upper bound on
parameter count in our experiments. We use the instruction-tuned variant
(\texttt{Llama-3.2-3B-Instruct}).

\paragraph{Qwen3-0.6B-Instruct~\cite{qwen3}.}
Qwen3-0.6B is a 0.6-billion-parameter model from Alibaba's Qwen3 series,
the smallest model in our evaluation. Its inclusion tests whether
\sysname's adaptive aggregation mechanism remains effective at the lower
end of the parameter scale, where gradient signals are weaker and the
policy's capacity to internalize group-specific preference distributions
is more constrained. We use the instruction-tuned variant
(\texttt{Qwen3-0.6B-Instruct}).

\paragraph{Design rationale.}
The three models were selected to span a representative range of open-weight
instruction-tuned LLMs (0.6B--3B parameters) from distinct model families
and training pipelines (Google DeepMind, Meta, Alibaba), ensuring that
empirical findings are not an artifact of any single architecture or
pretraining corpus. All models are used in 4-bit quantized form during PPO
to fit within the memory constraints of the A100 GPU cluster
(Appendix~\ref{appendix:setup}).

\section{Format Scoring and Final Reward Computation}
\label{appendix:format}

During PPO training, we observed that without enforcing an output format,
the policy often produced invalid responses, such as incorrect numbers of values,
extra tokens, or probability distributions that do not sum to one.
To address this, we introduce a format score $s_{\text{fmt}} \in [0,1]$
that is combined with the group-level reward $r^t_{g,j}$.

For \ac{dpa}, $s_{\text{fmt}}$ checks whether the model outputs the correct
number of values ($K$ comma-separated decimals), whether all values lie in $[0,1]$,
and whether they sum to one within a small tolerance.
For \ac{opa}, $s_{\text{fmt}}$ measures the fraction of valid, non-duplicate
option letters recovered out of the required $K$.

The final reward used for training is:
\begin{equation}
\label{eq:final-reward}
\tilde{r}^t_{g,j} \;=\; \omega \cdot r^t_{g,j} \;+\; (1 - \omega)
\cdot s_{\text{fmt},j}, \qquad \omega = 0.85,
\end{equation}
which combines the group-level reward with the format score.
Completely unparseable responses receive zero reward.

Before the PPO update, rewards are whitened per rollout batch and
clamped to $[-5, 5]$, following standard practice~\cite{schulman2017ppo}.

\section{Theoretical Justification of Adaptive Alpha Aggregation}
\label{app:theory}

We show two key properties of \sysname: (1) the aggregated reward is a valid scalar value that can be used in \ac{ppo}, and (2) the resulting update gives more weight to under-aligned groups based on their historical alignment rewards.

\subsection{Compatibility with the PPO Objective}

Standard \ac{ppo} maximizes an expected scalar reward $J(\theta)$
(Equation~\eqref{eq:ppo-standard}).
In \sysname, we replace the centralized reward with
$\mathrm{Agg}_{\boldsymbol{\alpha}^t}(\mathbf{r}^t_j)$ as defined in
Equation~\eqref{eq:adaptive-aggregation}.

Since $\mathrm{Agg}_{\boldsymbol{\alpha}^t}(\mathbf{r}^t_j)$ produces a bounded scalar reward at every iteration $t$, it can be directly used in the \ac{ppo} objective.
Although the aggregation is piecewise defined due to the threshold-based rule, \ac{ppo} only requires a scalar reward per trajectory element and does not rely on differentiability of the reward with respect to the policy parameters.
Therefore, the objective in Equation~\eqref{eq:ppo-federated} is well defined and stable under the same assumptions as prior \ac{rlhf} work~\cite{schulman2017ppo,ouyang2022training}.

Since all per-group rewards satisfy $r^t_{g,j} \in [0,1]$ and the adaptive weights $\alpha_g^t$ are positive and normalized via a softmax, the aggregated reward remains bounded for all $t$.

\subsection{Gradient Bias Toward Under-Aligned Groups}

Let $\theta$ denote the policy parameters.
\ac{ppo} optimizes $J(\theta) = \mathbb{E}[\mathrm{Agg}_{\boldsymbol{\alpha}^t}(\mathbf{r}^t_j)]$
and updates
$\theta \leftarrow \theta + \eta\,\nabla_\theta J(\theta)$.
Applying the chain rule:
\begin{equation}
\label{eq:app-grad-decomp}
    \nabla_\theta J(\theta)
    \;=\;
    \mathbb{E}\!\left[
        \sum_{g \in G_{\mathrm{train}}}
        \underbrace{
            \frac{\partial\;\mathrm{Agg}_{\boldsymbol{\alpha}^t}(\mathbf{r}^t_j)}
     {\partial\, r^t_{g,j}}
        }_{w_g^t}
        \;\nabla_\theta r^t_{g,j}(\theta)
    \right],
\end{equation}
where $w_g^t$ is the effective weight of group $g$ at iteration $t$.

\paragraph{Fair regime ($FI \ge \tau$).}
The aggregation reduces to a uniform average
(Equation~\eqref{eq:adaptive-aggregation}), so
$w_g^t = \tfrac{1}{|G_{\mathrm{train}}|}$ for all $g \in G_{\mathrm{train}}$.
Each group contributes equally to the update.

\paragraph{Unfair regime ($FI < \tau$).}
The aggregation switches to the adaptive log-sum-exp term.
The prefactor $1/|G_{\mathrm{train}}|$ in Equation~\eqref{eq:adaptive-aggregation}
appears in both numerator and denominator after differentiation and therefore cancels, yielding:
\begin{equation}
\label{eq:app-eff-weight}
    w_g^t
    \;=\;
    \frac{
        \alpha_g^t\,\exp\!\left(\alpha_g^t\, r^t_{g,j}\right)
    }{
        \displaystyle\sum_{g' \in G_{\mathrm{train}}}
        \exp\!\left(\alpha_{g'}^t\, r^t_{g',j}\right)
    }.
\end{equation}

Within each iteration, the adaptive weights $\alpha_g^t$ are treated as fixed with respect to $\theta$, since they are computed from historical alignment rewards.

Let $k = \arg\min_{g} h_g^{t-1}$ be the group with the lowest historical alignment reward.
Since $\alpha_g^t$ is computed via a softmax over $(1 - h_g^{t-1})/T$, the group with the lowest $h_g^{t-1}$ receives the largest weight $\alpha_g^t$:
\begin{equation}
    k \;=\; \arg\min_{g}\, h_g^{t-1}
    \quad\Longrightarrow\quad
    \alpha_k^t \;=\; \max_{g \in G_{\mathrm{train}}}\, \alpha_g^t.
\end{equation}

The effective weight $w_g^t$ depends on both $\alpha_g^t$ and $r^t_{g,j}$.
In practice, groups with larger adaptive weights tend to have greater influence on the update, especially when reward differences are not too large.
In particular, for groups with lower historical alignment rewards
($h_g^{t-1} < h_{g'}^{t-1}$), the reversed softmax produces
$\alpha_g^t > \alpha_{g'}^t$, increasing their influence in the aggregated update.

Equation~\eqref{eq:app-eff-weight} shows that when the spread of $\alpha_g^t$
(sharpened by $T{=}0.1$) is larger than the variation in rewards,
groups with larger $\alpha_g^t$ tend to receive larger effective weights $w_g^t$.
This behavior is common in practice but is not guaranteed for every rollout.

Substituting back, the update becomes
\begin{equation}
   \nabla_\theta J(\theta)
    \;=\;
    \mathbb{E}\!\left[
    \sum_{g \in G_{\mathrm{train}}}
    w_g^t\,\nabla_\theta r^t_{g,j}(\theta)
    \right],
\end{equation}
where the temperature $T{=}0.1$ sharpens the distribution and increases the emphasis on under-aligned groups. Thus, the update increases the reward of under-aligned groups while still keeping contributions from all groups, since $w_g^t > 0$ for all $g$. This differs from hard minimax approaches~\cite{maxmin,ramesh2024group}, which only update using the worst-performing group.

\paragraph{Self-correcting behavior.}
This creates a self-correcting loop: as a group's alignment improves, its history $h_g^t$ increases, which reduces $\alpha_g^t$ in later iterations and shifts focus to other under-aligned groups.

\paragraph{Relationship to \citet{park2024rlhf}.} They show that the $\mathrm{Agg}_\alpha$ family
(Equation~\eqref{eq:consensus-aggregation}) satisfies several fairness properties for a fixed scalar $\alpha$.
Our setting is different: we replace the fixed scalar with group-specific adaptive weights $\alpha_g^t$ based on historical alignment rewards.
This leads to the gradient form in Equation~\eqref{eq:app-grad-decomp}, which naturally gives more weight to under-aligned groups, a behavior that cannot be achieved with a single fixed $\alpha$.

\end{document}